%% file: main.tex
\documentclass{article}

\usepackage[preprint]{neurips_2026}
\makeatletter
\renewcommand{\@notice}{}
\makeatother
\usepackage[utf8]{inputenc} 
\usepackage[T1]{fontenc}    
\usepackage{url}            
\usepackage{graphicx}      
\usepackage{booktabs}       
\usepackage{amsmath}        
\usepackage{amsfonts}       
\usepackage{nicefrac}       
\usepackage{microtype}      
\usepackage{xcolor}         
\usepackage{algorithm}
\usepackage{algpseudocode}
\usepackage{float}
\usepackage{multirow}
\usepackage{tikz}
\usetikzlibrary{arrows.meta}
\usepackage{hyperref}       

\newcommand{\method}{Organon}
\newcommand{\flashmodel}{DeepSeek-V4.1-Flash}
\newcommand{\promodel}{DeepSeek-V4-Pro}
\title{Organon : From Evolutionary Search to Recursive Self-Improving Agents for Scientific Discovery}
\author{%
  Cheikh Ahmed\textsuperscript{1} \quad
  Zhiyu Zhang\textsuperscript{2} \quad
  Kun Mao\textsuperscript{2} \quad
  Yong Zhang\textsuperscript{1} \quad
  Zirui Zhou\textsuperscript{1} \\
  \normalfont\textsuperscript{1}Huawei Technologies Canada \quad
  \textsuperscript{2}Huawei Technologies
}

\begin{document}

\maketitle

\begin{abstract}
Scientific discovery does not follow a single fixed methodology: research methods vary across domains and evolve through experimentation as investigations unfold. Although LLM-guided evolutionary search has advanced automated discovery, existing frameworks typically rely on fixed search procedures or restrict adaptation to predefined components, limiting their ability to address diverse research problems and adapt to challenges that emerge at runtime. We introduce \method{}, a recursively self-improving agent for scientific discovery. During execution, a single persistent agent analyzes its accumulated experience and autonomously revises its own search strategies, population management, diagnostic tools, memory, and worker coordination. These revisions take effect at runtime, allowing the agent to continuously adapt its approach as research progresses. Importantly, the agent can also diagnose and correct flaws in its own adaptation procedures, enabling recursive improvement of the mechanisms that guide its future self-improvement. We evaluate \method{} across 21 computational discovery tasks spanning algorithm design, GPU kernel optimization, mathematical construction, and systems optimization. On ALE-Bench Lite, \method{} achieves the best performance among competing search frameworks on all 10 tasks, improving mean held-out performance by 20.5\% over EvoX. On KernelBench, it achieves the best results on three of four workloads, with speedups of up to $1.67\times$ over the strongest competing method. Analysis of the search trajectories further reveals how the agent develops task-specific search strategies, builds new diagnostic capabilities, and corrects or reverses its own earlier revisions. We implement \method{} using DeepSeek Harness and Cordis's spatiotemporal composability. Code is available at \url{https://github.com/cheikh025/Organon}.

\end{abstract}

\begin{figure}[h]
    \centering
    \includegraphics[width=\linewidth]{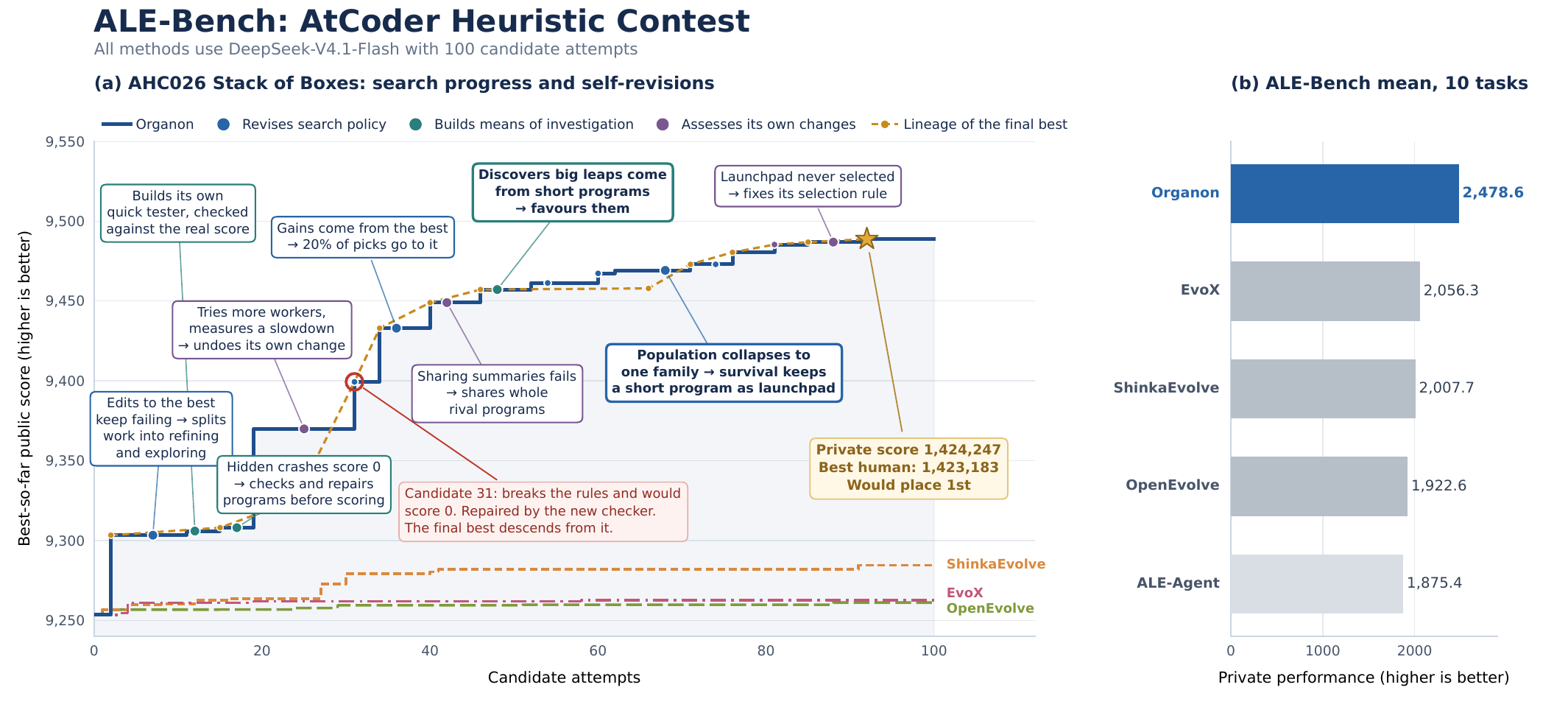}
    \caption{\method{} on ALE-Bench. (a) On AHC026, markers show the agent revising its own search; the gold dashed line traces the final best program, which would have placed first. (b) Mean private performance over the 10 ALE-Bench tasks.}
    \label{fig:teaser}
\end{figure}

\input{sections/introduction}

\input{sections/preliminaries}
\input{sections/methodology}

\input{sections/experiments}
\input{sections/anatomy}

\input{sections/related_work}
\input{sections/conclusion}

\bibliographystyle{plainnat}
\bibliography{references}

\clearpage
\appendix
\input{sections/appendix_anatomy}

\end{document}

%% file: sections/introduction.tex
\section{Introduction}
\label{sec:introduction}

Scientific progress depends on how competing hypotheses are generated, tested, and refined, and on how the methodology of inquiry itself evolves in light of empirical evidence. Large language models (LLMs) increasingly automate key stages of this process through autonomous reasoning, tool use, and experimentation \citep{aiscientist,Gottweis_2026}, while coding agents sustain long-horizon computational investigations \citep{karpathy2026autoresearch,lutke2026liquid,autokernel}. Throughout an investigation, these systems accumulate rich empirical histories spanning candidate solutions, execution profiles, and failed trials. Yet while existing agents steadily expand their knowledge of the target problem, the operational mechanisms governing how they conduct research remain largely static. A central challenge is enabling autonomous systems to translate accumulated experience into durable improvements in their own investigative process.

Evolutionary program search makes this challenge concrete by combining LLM-generated variation with computational evaluation to refine a population of executable candidates. FunSearch and AlphaEvolve established this paradigm for mathematical and algorithmic discovery \citep{romeraparedes2024funsearch,novikov2025alphaevolve}, with subsequent work advancing heuristic design, systems optimization, and reusable evolution frameworks \citep{liu2024eoh,ye2024reevo,cheng2025barbarians,lange2025shinkaevolve,assumpcao2025codeevolve}. In these systems, explicit mechanisms for parent selection, variation, memory organization, and resource allocation determine how accumulated evidence guides future experimentation. However, their effectiveness shifts as search progresses: a strategy that yields rapid early gains often concentrates subsequent effort within a narrow family of solutions, stalling long-horizon progress.

Recent evolutionary frameworks address this non-stationarity by adapting designated components of the search pipeline. ShinkaEvolve, PACEvolve, and AdaEvolve dynamically adjust population sampling, context management, exploration schedules, and resource allocation \citep{lange2025shinkaevolve,yan2026pacevolve,cemri2026adaevolve}. EvoX evolves strategies for parent selection, variation, and inspiration-context construction \citep{liu2026evox}, while Dream-RSI refines exploration policies through replay over recorded discovery trees while keeping its discovery agent fixed \citep{zheng2026dreamrsi}. Because these approaches confine adaptation to a predetermined subset of components and fixed update procedures, their adaptive scope is bounded at design time. When progress stalls due to bottlenecks outside those designated levers, or within the revision procedures themselves, the system cannot diagnose the underlying cause or restructure its operation.

Autonomous research agents offer complementary flexibility by interpreting evidence and directing experiments through step-by-step decisions, yet unconstrained in-context micro-decisions introduce an equally fundamental limitation. Over long horizons, an agent making ad-hoc local choices becomes absorbed in immediate task details and exhibits cognitive biases toward familiar or currently leading candidates. In single-incumbent workflows such as Autoresearch, this bias confines exploration to local, incremental refinement \citep{karpathy2026autoresearch,SwarmResearch}. Multi-agent frameworks such as CORAL broaden search through shared notes, reusable skills, and periodic heartbeat interventions, yet shared experience can also amplify leader-following and premature consensus, Configurable reflection and stagnation triggers provide opportunities for redirection without necessarily correcting the mechanisms sustaining unproductive search \citep{coral}. Even when SwarmResearch enforces centralized orchestration, controlled worker context, and isolated search branches, its orchestrator remains blind to its own decision biases, defaulting to near-greedy branch allocation and struggling to plan strategic experiments despite targeted prompting interventions \citep{SwarmResearch}. Together, these failure modes show why decision-level autonomy is insufficient: without diagnostic instruments to surface search pathologies and explicit, revisable mechanisms to govern selection, reflection, and coordination, step-by-step agent decisions might drift toward narrow, greedy behavior.

These limitations raise a fundamental question: can an autonomous agent's experience drive the continual self-improvement of how it discovers? We argue that it can, in the sense Francis Bacon described four centuries ago:

\begin{center}
    \begin{minipage}{0.85\linewidth}
        \centering
        {\large\itshape ``\ldots the art of discovery may advance as discoveries advance.''}\\[3pt]
        {\small  Francis Bacon, \emph{Novum Organum} (1620)}
    \end{minipage}
\end{center}

We introduce \method{}, a recursively self-improving evolutionary agent whose method of discovery advances with its discoveries. Its accumulated experience informs revisions to both the mechanisms governing search and the procedures governing its subsequent self-improvement. Rather than relying on ad-hoc micro-decisions or adapting a fixed subset of parameters, a single unified agent governs and modifies its entire operational stack, including the evolutionary loop, parent selection, candidate variation, population management, memory, tools, resource allocation, and worker coordination. Drawing on its accumulated population and search history, the agent decides when revision is warranted, which operational rules or capabilities to develop or modify, and how to evaluate their effects. Because each adopted revision becomes part of the active agent, self-improvement extends directly to the diagnostic tools, reusable skills, and decision procedures through which \method{} identifies its own limitations and evaluates subsequent changes. Revising a diagnostic tool, for example, can expose an unrecognized bottleneck or selection bias, prompting the agent to rewrite its governing mechanisms and redirect future capability development.

Realizing this form of self-improvement requires changing capabilities and governing rules during ongoing search without disrupting active investigations or severing the link between historical evidence and subsequent evaluation. We implement \method{} using DeepSeek Harness and Cordis's spatiotemporal composability \citep{deepseek2026harness,shi2026composability}. Reactive coeffects coordinate cross-component dependencies, while revertible effects isolate and manage each component's registered contributions. Together, they enable capabilities to be added, replaced, and withdrawn during live execution while preserving the accumulated population and search history, maintaining continuity between the experience that motivates a revision and the search used to assess its effects.

Our contributions are threefold:
\begin{enumerate}
    \item \textbf{Recursive self-improvement for discovery:} an agent that revises its own search, diagnostics, and revision procedures from accumulated experience.
    \item \textbf{State-preserving revision:} through spatiotemporal composability~\citep{shi2026composability}, capabilities are added, replaced, or withdrawn during search without losing the population or its history.
    \item \textbf{Evaluation and analysis:} strong results across 21 tasks, with an analysis of how the agent's capabilities evolved.
\end{enumerate}

%% file: sections/preliminaries.tex
\section{Preliminaries}
\label{sec:preliminaries}

We consider scientific discovery tasks in which candidate solutions are represented or generated by programs and assessed through computational evaluation. Let $\mathcal{X}$ denote the space of admissible programs, and let $f:\mathcal{X}\rightarrow\mathbb{R}$ be the official task evaluator, with larger scores indicating better performance. Starting from a baseline program $x_0\in\mathcal{X}$, the search seeks to maximize the best task score attained within a budget of $B$ candidate-generation attempts. If $x_1,\ldots,x_n$ are the candidates evaluated during the run, where $n\leq B$, the returned solution satisfies
\begin{equation}
    \hat{x}\in
    \arg\max_{x\in\{x_0,x_1,\ldots,x_n\}} f(x).
    \label{eq:discovery-objective}
\end{equation}
The official evaluator and candidate-generation budget remain fixed throughout the investigation.

Evolutionary search organizes candidate generation around a population of previously evaluated programs. Parent selection determines which candidates are developed, variation produces new candidates from the selected parents, and population management determines which evaluated candidates are retained for subsequent search. Together, these procedures govern how effort is distributed across competing approaches and how experimental feedback influences the development of new solutions.

The search accumulates a history of evaluated candidates, scores, evaluation feedback, and parent--offspring relationships. The active population provides candidates for further development, while the history records the outcomes of earlier search decisions. This evidence supports the assessment of candidate quality and the examination of search behavior, including which approaches are repeatedly pursued, which variations yield progress, and how retention shapes subsequent exploration. Accumulated experience therefore provides a basis for investigating how the procedures governing search can develop over the course of discovery.

%% file: sections/methodology.tex
\section{Method}
\label{sec:methodology}

\begin{figure}[!t]
    \centering
    \includegraphics[width=0.95\linewidth]{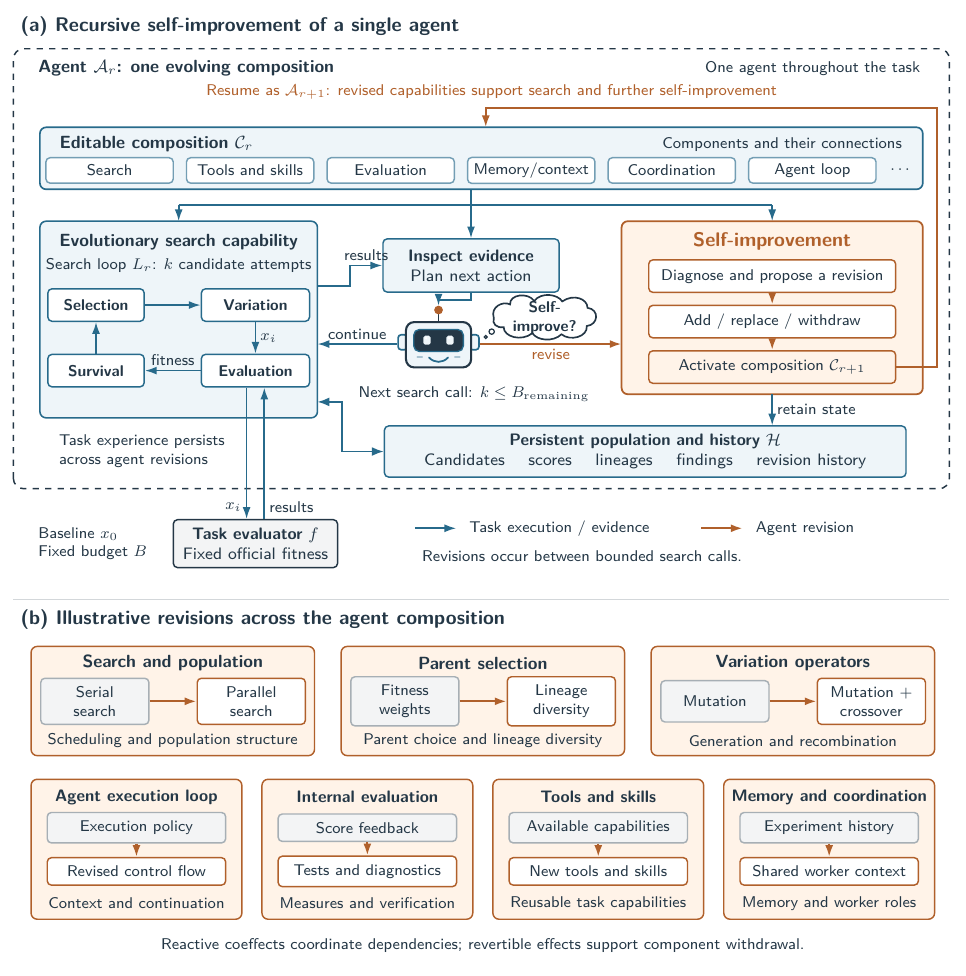}
    \caption{Overview of \method{}. (a) A single agent $\mathcal{A}_r$ conducts research through an editable composition $\mathcal{C}_r$, including its evolutionary search loop $L_r$, and revises itself between bounded search calls of $k \le B_{\text{remaining}}$ attempts; each adopted revision yields $\mathcal{A}_{r+1}$ with composition $\mathcal{C}_{r+1}$, while the population and history $\mathcal{H}$ persist. (b) Illustrative revisions across the agent composition.}
    \label{fig:evorsi_overview}
\end{figure}

\subsection{The Evolutionary Agent and Its Research Capabilities}
\label{subsec:agent_composition}

\subsubsection{Agent Composition and Persistent State}

\method{} conducts each investigation as a single continuing agent whose capabilities evolve with task experience while the problem definition $(x_0, f, B)$ remains fixed (Figure~\ref{fig:evorsi_overview}a). After $r \ge 0$ adopted capability revisions, let $\mathcal{A}_r$ denote the agent and $\mathcal{C}_r$ its active editable composition. The composition $\mathcal{C}_r$ encompasses all components governing the agent's operation and their connections:
\begin{equation}
    \mathcal{C}_r = \bigl(L_r,\, \mathcal{T}_r,\, \mathcal{M}_r,\, \mathcal{W}_r,\, \Phi_r\bigr),
    \label{eq:agent_composition}
\end{equation}
where $L_r$ is the evolutionary search capability, $\mathcal{T}_r$ denotes the available tools and skills (including internal diagnostics), $\mathcal{M}_r$ is the memory and context management, $\mathcal{W}_r$ specifies worker subagent coordination, and $\Phi_r$ is the agent's execution loop and instructions.

Alongside its editable composition $\mathcal{C}_r$, the agent maintains a persistent population and history $\mathcal{H}$ that accumulates across the entire run:
\begin{equation}
    \mathcal{H} = \bigl(\mathcal{P},\, \mathcal{R}_{\text{cand}},\, \mathcal{F}_{\text{work}},\, \mathcal{R}_{\text{rev}}\bigr),
    \label{eq:persistent_history}
\end{equation}
comprising the active population $\mathcal{P} \subseteq \mathcal{X}$, full candidate records $\mathcal{R}_{\text{cand}}$ (evaluated programs $x_i$, official scores $f(x_i)$, evaluation feedback, parent--offspring lineages, and survival status), worker findings $\mathcal{F}_{\text{work}}$, and the revision history $\mathcal{R}_{\text{rev}}$. While $\mathcal{C}_r$ defines how the agent currently researches and improves itself, $\mathcal{H}$ preserves the empirical evidence accumulated across all preceding search calls and agent revisions.

\subsubsection{Evolutionary Search as an Agent Capability}

Within $\mathcal{C}_r$, the evolutionary search capability is organized by a control loop $L_r$ coordinating four adaptable procedures:
\begin{equation}
    L_r = \bigl(\operatorname{Sel}_r,\, \operatorname{Var}_r,\, \operatorname{Eval}_r,\, \operatorname{Surv}_r\bigr),
    \label{eq:search_loop_components}
\end{equation}
corresponding to parent selection, candidate variation, internal evaluation protocol, and population survival. Search is initialized by evaluating the baseline program $x_0$ with $f$ and recording $(x_0, f(x_0))$ in $\mathcal{H}_0$ without consuming the attempt budget $B$. In the initial composition $\mathcal{C}_0$, $\operatorname{Sel}_0$ samples parents from the active population $\mathcal{P}$ using fitness-derived weights.

Each candidate attempt $x_i$ is a delegated development task carried out by a worker subagent under $\mathcal{W}_r$ in an isolated workspace initialized from the selected parent(s). \method{} supplies the worker's prompt, role, tools, skills, and relevant context from $\mathcal{H}$. The variation procedure $\operatorname{Var}_r$ organizes candidate development under this specification, allowing the agent to revise how workers are prepared, guided, and supported as the investigation progresses.

Once a worker produces a candidate $x_i$, the evaluation protocol $\operatorname{Eval}_r$ invokes the fixed task evaluator $f(x_i)$ to obtain the official fitness score and execution feedback, alongside any auxiliary tests or diagnostics in $\mathcal{T}_r$; development or evaluation failures are likewise recorded as attempt outcomes. The survival policy $\operatorname{Surv}_r$ then updates the active population $\mathcal{P}$ (initially retaining the highest-scoring candidates within population capacity) and appends the new record to $\mathcal{H}$:
\begin{equation}
    \begin{aligned}
        x_{\text{par}} &\sim \operatorname{Sel}_r(\mathcal{H}), \qquad
        &&x_i \sim \operatorname{Var}_r(x_{\text{par}}, \mathcal{H};\, \mathcal{W}_r, \mathcal{T}_r), \\
        y_i &= \operatorname{Eval}_r(x_i;\, f, \mathcal{T}_r), \qquad
        &&\mathcal{H} \leftarrow \operatorname{Surv}_r\bigl(\mathcal{H},\, (x_i, x_{\text{par}}, y_i)\bigr).
    \end{aligned}
    \label{eq:single_attempt}
\end{equation}
All four procedures in $L_r$ belong to the editable composition $\mathcal{C}_r$, whereas $f$ remains the fixed, authoritative scoring function. We realize this composition on DeepSeek Harness~\citep{deepseek2026harness}, using the spatiotemporal composability framework of~\citet{shi2026composability} detailed in Section~\ref{subsec:implementation_details}.

\subsection{Recursive Self-Improvement During Search}
\label{subsec:recursive_self_improvement}

\method{} couples two levels of evolution within a single task run: \emph{candidate evolution}, which updates the persistent search state $\mathcal{H}$ under the current composition $\mathcal{C}_r$, and \emph{capability evolution}, which revises the composition $\mathcal{C}_r \to \mathcal{C}_{r+1}$ using the evidence accumulated in $\mathcal{H}$.

\subsubsection{Research Horizons}

At each decision point, given a remaining attempt budget $B_{\text{remaining}} \in (0, B]$, the agent $\mathcal{A}_r$ selects a \emph{research horizon} $k \in \{1, \dots, B_{\text{remaining}}\}$ specifying how many candidate-generation attempts to allocate before reassessing the investigation. During the horizon, $\mathcal{C}_r$ is held fixed while the search loop $L_r$ executes up to $k$ candidate attempts according to Equation~\eqref{eq:single_attempt}:
\begin{equation}
    \mathcal{H}' = L_r\bigl(\mathcal{H};\, \mathcal{C}_r,\, f,\, k\bigr),
    \qquad
    B_{\text{remaining}}' = B_{\text{remaining}} - k.
    \label{eq:horizon_step}
\end{equation}
Each allocated candidate attempt consumes one unit of $B$, including attempts that fail during development or evaluation.

When the bounded search call returns, $\mathcal{A}_r$ inspects the updated evidence in $\mathcal{H}'$, including score progression, lineage concentration, evaluation feedback, and worker findings, and chooses whether to continue searching with the current composition $\mathcal{C}_r$, revise its capabilities to form $\mathcal{C}_{r+1}$, or conclude the investigation:
\begin{equation}
    \mathcal{C}_{r+1} =
    \begin{cases}
        \mathcal{C}_r, & \text{if } \mathcal{A}_r \text{ continues under the current composition}, \\[3pt]
        \operatorname{Revise}\bigl(\mathcal{C}_r,\, \mathcal{H}'\bigr), & \text{if } \mathcal{A}_r \text{ adopts a capability revision}.
    \end{cases}
    \label{eq:composition_transition}
\end{equation}
Several research horizons may therefore proceed under the same composition before a revision is adopted. Search ends when $B_{\text{remaining}} = 0$ or when the agent judges the investigation complete, returning the highest-scoring officially evaluated candidate $\hat{x} \in \arg\max_{x \in \mathcal{H}} f(x)$ (including $x_0$) and a final report.

\subsubsection{From Evidence to Capability}

When $\mathcal{A}_r$ elects to revise itself, it uses its reasoning loop $\Phi_r$, tools $\mathcal{T}_r$, and accumulated history $\mathcal{H}'$ to diagnose limitations in its current operation and construct a capability change (Figure~\ref{fig:evorsi_overview}b). A revision can introduce, modify, or withdraw a component, or alter how components interact across any part of $\mathcal{C}_r$: search scheduling and population structure ($L_r$), parent selection ($\operatorname{Sel}_r$), variation operators ($\operatorname{Var}_r$), internal evaluation and verification ($\operatorname{Eval}_r$), reusable tools and skills ($\mathcal{T}_r$), memory and worker coordination ($\mathcal{M}_r, \mathcal{W}_r$), or the agent execution loop ($\Phi_r$).

For illustration, consider a search that repeatedly expands one high-scoring lineage while its descendants yield diminishing returns. By inspecting $\mathcal{H}'$, the agent may first add a diagnostic tool to $\mathcal{T}_r$ that summarizes effort allocation and score gains across branches of the candidate ancestry graph. It can then revise the parent-selection policy $\operatorname{Sel}_r$ to weight lineages using these diagnostic summaries, redirecting subsequent attempts toward under-explored, promising branches.

Similarly, recurring compilation or runtime failures recorded in $\mathcal{H}'$ can motivate changes to candidate development. The agent may revise worker instructions in $\mathcal{W}_r$, add a static checking or repair skill to $\mathcal{T}_r$, or record known failure modes and remedies in memory $\mathcal{M}_r$ for future worker preparation. Because diagnostics and assessment procedures are themselves part of $\mathcal{C}_r$, refining how $\mathcal{H}'$ is analyzed directly improves the agent's capacity to propose and evaluate subsequent self-improvements.

\subsubsection{Self-Improvement Through Composition}

A capability revision transforms $\mathcal{C}_r$ into $\mathcal{C}_{r+1}$ through integration into the live runtime context. Let $\Gamma_r$ denote the runtime context realizing $\mathcal{C}_r$. A component $p$ participates in $\Gamma_r$ when its declared dependencies (coeffects) are satisfied. Activating $p$ applies an effect $e_p$ that yields a revised context $\Gamma_{r+1}$ together with a tracked reversion $u_p$:
\begin{equation}
    e_p(\Gamma_r) = (\Gamma_{r+1},\, u_p),
    \qquad
    u_p(\Gamma_{r+1}) \simeq \Gamma_r.
    \label{eq:component_revision}
\end{equation}
Here $\simeq$ denotes equivalence of the active context after withdrawing $p$'s registered contributions with all other components held fixed~\citep{shi2026composability}. Reactive coeffects automatically coordinate component activation as dependencies change, while revertible effects allow any adopted revision to be cleanly replaced or withdrawn if subsequent horizons show that it is ineffective.

Returning to the lineage example, let $\Gamma_r$ include the lineage diagnostic and let component $p$ register a revised parent-selection policy $\pi_p$. Activating $p$ updates the selector bound to the search loop while preserving reversibility:
\begin{equation}
    \operatorname{Sel}(\Gamma_{r+1}) = \pi_p,
    \qquad
    \operatorname{Sel}\bigl(u_p(\Gamma_{r+1})\bigr) = \operatorname{Sel}(\Gamma_r).
    \label{eq:selector_revision}
\end{equation}
Revisions take effect between research horizons. Once $\Gamma_{r+1}$ (realizing $\mathcal{C}_{r+1}$) is active, the agent resumes as $\mathcal{A}_{r+1}$, launches the next bounded search horizon $L_{r+1}(\mathcal{H}';\, \mathcal{C}_{r+1}, f, k')$, and evaluates from the resulting outcomes whether to retain, refine, or withdraw $p$. Crucially, while $\mathcal{C}_r \to \mathcal{C}_{r+1}$ evolves across revisions, the persistent history $\mathcal{H}'$, remaining budget $B_{\text{remaining}}'$, and official evaluator $f$ remain invariant across composition changes.

\subsection{Implementation Details}
\label{subsec:implementation_details}

We implement \method{} on DeepSeek Harness~\citep{deepseek2026harness}, whose dynamically composed runtime supplies model access, instructions, tools, session state, and the agent's execution loop through replaceable components. Within this runtime, we add an evolutionary search service exposing five replaceable provider interfaces: the evolutionary loop, parent selection, candidate variation, evaluation, and population survival. These interfaces are part of the agent's active composition alongside its skills, memory, worker subagents, and other capabilities. The initial providers realize the search described in Section~\ref{subsec:agent_composition}. Each call to the service uses the active provider for every operation, including the evolutionary loop, so a provider change governs subsequent candidate search.

The agent revises these providers by defining a dynamic component with declared service dependencies and activating it in its session. A component registers its contribution through a tracked effect; the most recently registered provider for a given operation becomes active. Deactivating the component withdraws its registration and restores the preceding available provider. The agent can also activate a revised version of a component. These changes occur between bounded calls to the evolutionary loop, after the agent has examined the results of the preceding call and decided how to proceed.

The search state persists across these calls. Candidate records retain parent identifiers, fitness scores, and survival status, while evaluation artifacts provide feedback for subsequent decisions. The task's evaluator supplies the official fitness $f$, and a new candidate consumes one unit of the fixed budget $B$ when it is allocated; the initial program $x_0$ consumes none. The agent chooses how many candidates to attempt in each call, can inspect the remaining budget and active providers, and may use additional diagnostics to guide further search or self-improvement.

%% file: sections/experiments.tex
\section{Experiments}
\label{sec:experiments}

We evaluate the quality of programs discovered by
\method{} across algorithm design, kernel engineering,
mathematical construction, and system optimization.
The experiments examine performance under a common
candidate-generation allowance and characterize how
the resulting advantages vary across task families.

\subsection{Experimental Setup}
\label{subsec:experimental_setup}

\paragraph{Tasks and comparison methods.}
We consider 21 tasks across four categories.
Algorithm design comprises the ten ALE-Bench Lite
tasks~\citep{imajuku2025alebench}. Kernel engineering
comprises four KernelBench workloads: VGG16, LayerNorm,
ConvDiv, and ConvMax~\citep{ouyang2025kernelbench}.
Mathematical construction includes packing 26 circles
in a unit square, minimizing third-order autocorrelation,
and the Heilbronn triangle problem with 11 points in
an equilateral triangle. System optimization includes
Model Placement (PRISM), Expert Parallelism Load
Balancer (EPLB), Column Reordering for Prefix Caching
(LLM-SQL), and Multi-Cloud Data Transfer Optimization
(Cloudcast)~\citep{cheng2025barbarians}. Each task
provides a starting program and a task-specific
evaluator.

We compare \method{} with
OpenEvolve~\citep{sharma2025openevolve},
ShinkaEvolve~\citep{lange2025shinkaevolve},
and EvoX~\citep{liu2026evox}. These methods are run
through SkyDiscover~\citep{liu2026skydiscover},
using its EvoX implementation and external backends
for OpenEvolve and ShinkaEvolve. All four methods
use \flashmodel{} with high reasoning effort
and a maximum of $B=100$ candidate-generation
attempts per task.\footnote{On AHC016, AHC046, and third-order autocorrelation, the agent routed 38 of 300 candidates to \promodel{}, which ranks below \flashmodel{} on independent evaluations~\citep{artificialanalysis2026deepseek}.} For a fair comparison, worker subagents operate in a sandbox without shell or web access: each can only edit the source of its own candidate and cannot compile, execute, or test it. This matches the comparison methods, whose language model likewise proposes code without running it, and prevents a worker from substituting local trial-and-error for evaluated candidate attempts. Execution-based checks therefore arise only when the agent introduces them as part of its own search capabilities (Appendix~\ref{subsec:app_anatomy_variation}). The starting program is evaluated separately and does not consume this budget. Evaluation of the starting
program is excluded from this allowance.
The comparison therefore aligns the language-model
backbone and maximum candidate count.

\paragraph{Evaluation and references.}
On ALE-Bench, search and program selection use
50 public test cases. We report the mean private
performance over three evaluations of each selected
program and average these values across the ten
tasks for the aggregate result. Kernel results
are speedups relative to PyTorch eager, with
correctness checks and runtime measurements
conducted through Modal. Mathematical results
are best-found objective values. System results
are means over three evaluations of each selected
program.

We report ALE-Agent, which supplies the ALE-Bench
starting programs, as an initialization reference.
Mathematical references are computed from
AlphaEvolve's released
constructions~\mbox{\citep{googledeepmind2025alphaevolve_results}}.
System references are the human-best values reported
in EvoX~\citep{liu2026evox}. These references provide
context for attained solution quality; the four
re-run search methods form the comparison under
the common candidate allowance.

\subsection{ALE-Bench Results}
\label{subsec:ale_bench_results}

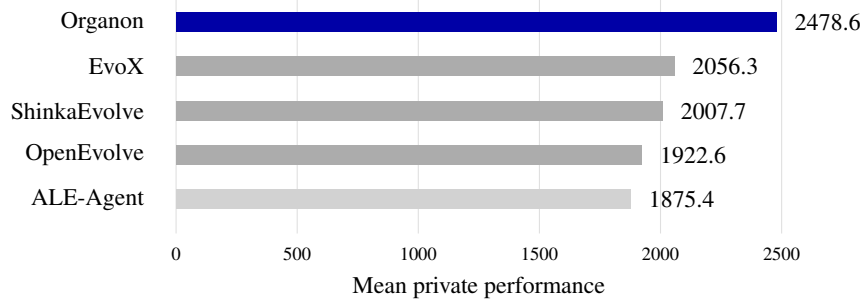
\begin{figure}[!htbp]
    \centering
    \input{figures/ale_mean_performance.tikz}
    \caption{Mean private performance across the ten
    ALE-Bench Lite tasks. Higher is better. ALE-Agent
    is an initialization reference.}
    \label{fig:ale_mean_rating}
\end{figure}

ALE-Bench evaluates whether programs selected using
public cases retain their advantage on private cases.
Figure~\ref{fig:ale_mean_rating} reports mean private
performance, and Table~\ref{tab:ale_ratings} shows
the task-level results.

\begin{table}[!htbp]
\centering
\caption{Private performance on ALE-Bench Lite.
Task values are means over three evaluations of
the selected program; the final row averages across
tasks. Higher is better. Bold marks the largest
reported value among the four compared search
methods. ALE-Agent is an initialization reference.}
\label{tab:ale_ratings}
\resizebox{\linewidth}{!}{%
\begin{tabular}{@{}llrrrrr@{}}
\toprule
\textbf{ID} & \textbf{Task} & \textbf{ALE-Agent}
& \textbf{OpenEvolve} & \textbf{ShinkaEvolve}
& \textbf{EvoX} & \textbf{\method{}} \\
\midrule
\texttt{AHC008} & Territory
& 1186.0 & 1257.0 & 1247.0 & 1265.0 & \textbf{1323.0} \\
\texttt{AHC011} & Sliding Tree Puzzle
& 1652.0 & 1675.0 & 1665.0 & 1652.0 & \textbf{1856.3} \\
\texttt{AHC015} & Halloween Candy
& 2405.0 & 2411.0 & 2481.7 & 2638.7 & \textbf{3043.0} \\
\texttt{AHC016} & Graphorean
& 1450.7 & 1496.0 & 1547.0 & 2120.3 & \textbf{2943.0} \\
\texttt{AHC024} & Topological Map
& 1977.0 & 2027.0 & 2254.3 & 2231.7 & \textbf{2262.7} \\
\texttt{AHC025} & Balancing by Balance
& 1370.7 & 1372.0 & 1901.0 & 1421.0 & \textbf{2017.3} \\
\texttt{AHC026} & Stack of Boxes
& 1981.3 & 1977.7 & 2083.0 & 1983.0 & \textbf{3451.0} \\
\texttt{AHC027} & Recurring Cleaning Route
& 1731.3 & 1869.0 & 1862.0 & 2015.0 & \textbf{2188.0} \\
\texttt{AHC039} & Purse Seine Fishing
& 2847.3 & 2947.3 & 2880.0 & 2974.0 & \textbf{3140.0} \\
\texttt{AHC046} & Skating with Blocks
& 2153.0 & 2194.0 & 2155.7 & 2262.0 & \textbf{2562.0} \\
\midrule
\multicolumn{2}{l}{Mean performance $\uparrow$}
& 1875.4 & 1922.6 & 2007.7 & 2056.3 & \textbf{2478.6} \\
\bottomrule
\end{tabular}%
}
\end{table}

\method{} achieves the highest mean private performance
of 2478.6, exceeding EvoX, the strongest comparison
method on average, by 20.5\% and the ALE-Agent reference
by 32.2\%. It obtains the highest reported score on
all ten tasks, with gains varying in magnitude across
problems. Since search and program selection use
public cases, these results establish that the
discovered improvements extend to held-out cases
within the evaluated tasks.

\subsection{KernelBench Results}
\label{subsec:kernelbench_results}

KernelBench tests whether discovered programs yield
faster implementations under computational correctness
checks. Table~\ref{tab:kernelbench_results} reports
speedups relative to PyTorch eager on four selected
workloads. Correctness checks and timing are conducted
through Modal using the KernelBench evaluation
procedure.\footnote{\url{https://github.com/ScalingIntelligence/KernelBench/blob/main/EVAL.md}}

\begin{table}[!htbp]
\centering
\caption{KernelBench speedups relative to PyTorch
eager. Higher is better. Bold marks the largest
reported speedup on each workload.}
\label{tab:kernelbench_results}
\small
\begin{tabular}{@{}lrrrr@{}}
\toprule
\textbf{Task} & \textbf{OpenEvolve}
& \textbf{ShinkaEvolve} & \textbf{EvoX}
& \textbf{\method{}} \\
\midrule
VGG16 & $2.10\times$ & $1.96\times$
& $1.89\times$ & $\mathbf{3.34\times}$ \\
LayerNorm & $16.13\times$ & $17.44\times$
& $16.56\times$ & $\mathbf{18.70\times}$ \\
ConvDiv & $1.90\times$ & $3.34\times$
& $3.26\times$ & $\mathbf{5.58\times}$ \\
ConvMax & $2.16\times$ & $2.13\times$
& $\mathbf{2.75\times}$ & $2.29\times$ \\
\bottomrule
\end{tabular}
\end{table}
\method{} achieves the highest speedup on three
of the four workloads, with execution-time reductions
of up to 40.1\% relative to the strongest comparator
on the same workload. The remaining workload favors
EvoX, indicating that the comparative advantage
depends on the optimization problem. Overall,
the results demonstrate strong kernel optimization
performance under the common candidate allowance,
extending the method's applicability to
runtime-sensitive program discovery.

\subsection{Mathematical Construction Results}
\label{subsec:math_results}

The mathematical tasks evaluate the objective values
of the constructions produced by search.
Table~\ref{tab:math_results} reports the attained
values alongside references computed from released
AlphaEvolve constructions.

\begin{table}[!htbp]
\centering
\caption{Mathematical construction results: sum
of radii for circle packing, $C_3$ for autocorrelation,
and minimum triangle area normalized by the
containing triangle's area for Heilbronn.
Arrows indicate the direction of improvement;
bold marks the best reported value among the
four compared search methods, including ties.
AlphaEvolve references are computed from released
constructions~\citep{googledeepmind2025alphaevolve_results}.}
\label{tab:math_results}
\resizebox{\linewidth}{!}{%
\begin{tabular}{@{}lrrrrr@{}}
\toprule
\textbf{Task} & \textbf{AlphaEvolve ref.}
& \textbf{OpenEvolve} & \textbf{ShinkaEvolve}
& \textbf{EvoX} & \textbf{\method{}} \\
\midrule
Circle packing, $n=26$ $\uparrow$
& 2.635863 & 2.630188 & 2.635983
& \textbf{2.635987} & 2.635314 \\
Third-order autocorrelation $\downarrow$
& 1.4556428 & 1.4566544 & 1.4575646
& 1.4570839 & \textbf{1.4562828} \\
Heilbronn triangle $\uparrow$
& 0.0365298899 & 0.0365295964
& \textbf{0.0365298899}
& \textbf{0.0365298899} & 0.0365035354 \\
\bottomrule
\end{tabular}%
}
\end{table}

\method{} obtains the best third-order autocorrelation result among the compared methods and comes closest to the AlphaEvolve construction reference. On the geometric tasks, it produces constructions close to the reference values, but EvoX and ShinkaEvolve attain stronger results. The mathematical results therefore distinguish successful optimization on autocorrelation from a remaining gap on circle packing and Heilbronn.

The geometric shortfall may reflect difficulty progressing beyond strong configurations within the available budget. Once search reaches such a configuration, further refinement can yield diminishing gains, while alternative configurations may be difficult to discover. EvoX likewise reports stagnation associated with strong local optima in its mathematical experiments~\citep{liu2026evox}. This offers a plausible interpretation of the geometric results, while the autocorrelation win shows that the limitation does not extend uniformly across the mathematical tasks.

\subsection{System Optimization Results}
\label{subsec:system_results}

System optimization assesses the resulting programs
against application-specific objectives.
PRISM, EPLB, and LLM-SQL are maximization tasks;
Cloudcast minimizes transfer cost.
Table~\ref{tab:system_results} separates comparison
among the search methods from comparison with
the cited human-best references.

\begin{table}[!htbp]
\centering
\caption{System-optimization objectives, averaged
over three evaluations of each selected program.
Arrows indicate the direction of improvement;
bold marks the best reported mean among the
four compared search methods, including ties.
Human-best references are from Table~3 of EvoX~\citep{liu2026evox}.}
\label{tab:system_results}
\small
\begin{tabular}{@{}lrrrrr@{}}
\toprule
\textbf{Task} & \textbf{Human best}
& \textbf{OpenEvolve} & \textbf{ShinkaEvolve}
& \textbf{EvoX} & \textbf{\method{}} \\
\midrule
PRISM $\uparrow$
& 21.89 & \textbf{26.2560} & \textbf{26.2560}
& \textbf{26.2560} & \textbf{26.2560} \\
EPLB $\uparrow$
& 0.1265 & 0.1461 & 0.1456 & 0.1471
& \textbf{0.1473} \\
LLM-SQL $\uparrow$
& 0.6920 & 0.7284 & 0.7282 & 0.7194
& \textbf{0.7367} \\
Cloudcast $\downarrow$
& 626.24 & 618.0945 & 619.9607 & 618.0876
& \textbf{618.0596} \\
\bottomrule
\end{tabular}
\end{table}

Every compared method improves on the cited
human-best references across all four tasks.
On PRISM, all methods attain 26.2560, approximately
19.9\% above the reference. This shared outcome
demonstrates the value of automated search relative
to that reference while offering limited
discrimination among the evaluated methods.

The largest comparative margin for \method{}
occurs on LLM-SQL. Its objective of 0.7367 exceeds
OpenEvolve's 0.7284 by 0.0083, or 1.14\%.
EPLB and Cloudcast show much closer results:
the respective improvements over EvoX are 0.0002
and 0.0280, corresponding to approximately
0.136\% and 0.0045\%. These are the best reported
means, with their small margins making variability
an important part of interpreting the ranking.
The system results thus identify LLM-SQL as the
clearest comparative gain, alongside a shared
PRISM outcome and closely spaced EPLB and Cloudcast
results.

%% file: figures/ale_mean_performance.tikz
\begin{tikzpicture}[x=0.0032cm,y=0.65cm,font=\small]
  \foreach \rating in {0,500,1000,1500,2000,2500} {
    \draw[gray!25] (\rating,0.5) -- (\rating,5.3);
    \node[font=\scriptsize,anchor=north] at (\rating,0.4) {\rating};
  }

  \fill[blue!65!black] (0,4.6) rectangle (2478.6,5.0);
  \fill[gray!65] (0,3.7) rectangle (2056.3,4.1);
  \fill[gray!65] (0,2.8) rectangle (2007.7,3.2);
  \fill[gray!65] (0,1.9) rectangle (1922.6,2.3);
  \fill[gray!35] (0,1.0) rectangle (1875.4,1.4);

  \node[anchor=east,xshift=-0.3cm] at (0,4.8) {\method{}};
  \node[anchor=east,xshift=-0.3cm] at (0,3.9) {EvoX};
  \node[anchor=east,xshift=-0.3cm] at (0,3.0) {ShinkaEvolve};
  \node[anchor=east,xshift=-0.3cm] at (0,2.1) {OpenEvolve};
  \node[anchor=east,xshift=-0.3cm] at (0,1.2) {ALE-Agent};

  \node[anchor=west,xshift=0.12cm] at (2478.6,4.8) {2478.6};
  \node[anchor=west,xshift=0.12cm] at (2056.3,3.9) {2056.3};
  \node[anchor=west,xshift=0.12cm] at (2007.7,3.0) {2007.7};
  \node[anchor=west,xshift=0.12cm] at (1922.6,2.1) {1922.6};
  \node[anchor=west,xshift=0.12cm] at (1875.4,1.2) {1875.4};

  \node at (1250,-0.6) {Mean private performance};
\end{tikzpicture}

%% file: sections/anatomy.tex
\section{Anatomy of Agent Self-Improvement}
\label{sec:anatomy}
\begin{figure}[H]
    \centering
    \includegraphics[width=\linewidth]{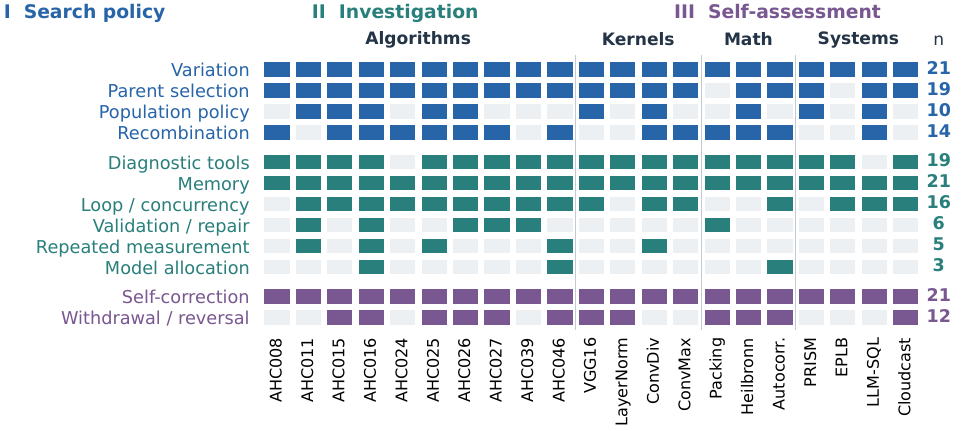}
    \caption{Recorded capability changes across 21 runs. Colored cells indicate changes exercised during search; gray cells indicate none recorded. Colors group the three perspectives below; $n$ counts runs. Variation, memory, and self-correction occur in every run.}
    \label{fig:anatomy_synthesis}
\end{figure}

\method{}'s search trajectories reveal how accumulated experience reshaped the agent's research process, from the policies directing search to the capabilities used to diagnose limitations and assess revisions. The agent revised its composition on all 21 tasks, first after a median of four candidate attempts (range 1--8), with 2--22 activated revisions per run (median 11), each motivated by measured evidence (Table~\ref{tab:app_anatomy_diagnoses}). On seven tasks, no candidate generated under the initial composition $\mathcal{C}_0$ improved on $x_0$; on 14 of 21, at least 70\% of the total improvement followed the first revision (Table~\ref{tab:app_anatomy_timing}). Figure~\ref{fig:anatomy_synthesis} shows that these changes were broad and task-specific: no two runs modified the same subset of capabilities (Table~\ref{tab:app_anatomy_matrix}). Three connected perspectives organize the analysis: revising the search policy, developing the means of investigation, and assessing the agent's own changes.

\paragraph{I. Revising the search policy.}
\method{} developed a task-specific search policy through the co-evolution of variation, selection, and survival. Evidence that unconstrained rewrites disrupted effective mechanisms prompted a shift from generic mutation to portfolios separating bounded refinement from structural redesign, with specialized operators targeting runtime margins, isolated design choices, and inactive mechanisms (Table~\ref{tab:app_anatomy_roles}). Recombination expanded inheritance across five granularities on 14 tasks, including individual contributions, measured gains, and design rationales. The allocation and scale of variation followed measured yield: AHC024 raised crossover from one-third to two-thirds of attempts after observing mean parent-relative gains of $+4.3$ versus $+0.5$ for other operators, while AHC008 favored 20--140-line edits, which improved their parents in 24 of 61 attempts, compared with 1 of 11 for larger revisions. Selection evolved in parallel, combining rank-, tournament-, and elite-based sampling with load-aware weighting (AHC025), recency penalties (AHC027), and stagnation-triggered broadening (AHC008, ConvDiv) to balance exploitation against redundant concentration. Crucially, variation's growing specialization changed which candidates were valuable to select and preserve. In AHC026, substantial redesign gains originated exclusively from parents below approximately $27$\,KB, motivating size-aware selection; discovering that top-$k$ survival had eliminated these candidates then prompted preservation of the strongest compact program. More broadly, survival extended beyond fitness-based retention to safeguard behavioral niches (LLM-SQL), distinct ancestries (Heilbronn), repairable failures (VGG16), and successful archived candidates.
\paragraph{II. Developing the means of investigation.}
The agent also expanded the diagnostic instruments, execution workflows, and memory governing subsequent research. On 19 tasks, custom measurement tools determined what to revise next: harness analysis in LayerNorm traced seven of eight initial failures to an indented Triton import rule (unlocking a $15.98\times$ kernel); per-case profiles in AHC011 showed the baseline never completed a tree; repeated-mean ranking (5 tasks) mitigated $\pm 5\%$ evaluation noise in ConvDiv (raising speedup from $3.29\times$ to $5.93\times$); and feasibility and bandwidth probes in PRISM and LayerNorm established objective saturation, redirecting search toward simplicity and cache reuse. Workflow revisions (16 tasks) paired concurrent or streaming generation with serialized evaluation, inserted pre-evaluation compilation and regression repair (6 tasks), tuned concurrency from throughput measurements (AHC026 reverted from eight to five workers after measuring $0.205$ vs.\ $0.269$ candidates/min), and lengthened assessment horizons as compositions stabilized (19 of 21 tasks). Meanwhile, memory (21 tasks; Appendix~\ref{subsec:app_anatomy_memory}) evolved from passive digests into active control driving topic-based exploration (AHC008) and gain-merge assignments (AHC046). In circle packing, diagnosing that time-dependent constructors failed to reproduce the frontier from code alone led the agent to store and verify the best 26-circle arrangement as numerical data, ending a 17-attempt plateau ($2.6287 \to 2.6353$).

\paragraph{III. Assessing and revising its own changes.}
The same empirical scrutiny extended to the agent's revised capabilities. Assessment addressed both whether a change operated as intended and whether its continued use was supported by subsequent outcomes. Component histories exposed discrepancies that task scores alone could not reveal: in AHC025, an inverted survival rule retained the five weakest candidates, and correcting it revived a culled candidate from which the final winner descended (Table~\ref{tab:app_anatomy_diagnoses}). This scrutiny also reached the instruments guiding further revision. Correcting AHC011's completion check exposed a twelve-tree regression and changed subsequent worker guidance, revising the evidence on which later interventions depended. Withdrawal complemented correction on 12 tasks. AHC016 abandoned alternative-model routing after ten candidates yielded no gain at more than twice the generation time; Cloudcast restored a prior component version while retaining the population and diagnostics accumulated under its replacement. State-preserving reversal allowed the agent to reject an operational hypothesis without discarding the experience acquired under it.

%% file: sections/related_work.tex
\section{Related Work}
\label{sec:related-work}

\paragraph{LLM-guided evolutionary search.}
LLM-guided evolution combines code generation with empirical evaluation and reuse of earlier candidates. ELM and FunSearch established this approach through language-model variation and populations of evaluated programs~\citep{lehman2022elm,romeraparedes2024funsearch}. AlphaEvolve~\citep{novikov2025alphaevolve} extends it to larger programs and richer feedback, including co-evolved prompt instructions and context. Later methods improve candidate sampling and model selection~\citep{lange2025shinkaevolve}, or represent search history through semantic program changes~\citep{jiang2026deltaevolve}. Search adaptation also includes progress-based trajectory management~\citep{yan2026pacevolve} and control of exploration and resource allocation~\citep{cemri2026adaevolve}. EvoX~\citep{liu2026evox} directly evolves strategies for parent selection, variation-operator choice, and inspiration context, using stagnation and population history to guide revisions. CORAL and SwarmResearch introduce
autonomous multi-agent exploration and coordinated search
~\citep{coral,SwarmResearch}. Dream-RSI~\citep{zheng2026dreamrsi} evaluates exploration policies through replay of recorded discovery trees before redeploying them with a fixed discovery agent. Both use discovery experience to improve subsequent search decisions.

\paragraph{Self-evolving agents.}
Self-evolving agents revise the mechanisms governing task execution. Harness optimization ranges from prompt-level workflow revision~\citep{lee2026rhi} to changes in executable harnesses, context management, tools, and memory~\citep{lee2026metaharness,lin2026ahe,liu2026solpi}. Complementary work addresses the search over agent variants: SIFT~\citep{fu2026sift} uses inexpensive pairwise judgments alongside benchmark feedback in asynchronous tree search, while RRSI~\citep{xia2026rrsi} regularizes proposals and selection to discourage benchmark-specific edits, unstable gains, and unjustified resource growth. Recursive approaches make agent code part of the improvement process. STOP~\citep{zelikman2023stop} applies an improver to its own source, and the Darwin G\"odel Machine~\citep{zhang2025dgm} evolves an archive of self-modifying coding agents. Hyperagents~\citep{zhang2026hyperagents} make both task execution and the mechanism generating future modifications editable. AIDE$^2$~\citep{srikanth2026aide2} uses an outer research agent to rewrite an inner optimization agent, selecting revisions through downstream research performance under fixed task budgets.

%% file: sections/conclusion.tex
\section{Conclusion}
\label{sec:conclusion}

We introduced \method{}, a recursively self-improving evolutionary agent that develops its research capabilities during discovery. A single persistent agent revises its search policies, its diagnostic instruments, and the procedures that govern its further revision, while retaining its population and accumulated experience. Across 21 tasks, \method{} achieves strong results in algorithm design, kernel engineering, and system optimization and remains competitive in mathematical construction, and its trajectories show accumulated evidence becoming task-specific strategies, tools, and memory. Future work should isolate the contribution of individual revisions through repeated runs and controlled ablations, and carry capabilities developed on one task into the next, so that self-improvement compounds across problems as well as within them.

%% file: sections/appendix_anatomy.tex
\section{Detailed Anatomy of Agent Self-Improvement}
\label{sec:appendix_anatomy}

This appendix provides the evidence for Section~\ref{sec:anatomy}. Section~\ref{subsec:app_anatomy_protocol} describes the records and how revisions are identified. Section~\ref{subsec:app_anatomy_timing} relates the timing of self-modification to search progress, and Section~\ref{subsec:app_anatomy_inventory} gives a per-run inventory. Sections~\ref{subsec:app_anatomy_variation}--\ref{subsec:app_anatomy_measurement} examine each capability category, Section~\ref{subsec:app_anatomy_correction} documents self-correction and withdrawal, and Section~\ref{subsec:app_anatomy_trajectories} presents representative trajectories.

\subsection{Records and Analysis Protocol}
\label{subsec:app_anatomy_protocol}

For each of the 21 tasks we analyze one complete run of \method{}. The records comprise the agent's complete session, every component it defined together with the diagnosis it recorded when defining it, the results of every activation and withdrawal, every search call with the number of candidate attempts it consumed, the population record of each candidate (score, parent, survival status, and any role or donor annotation), evaluation artifacts, the tools and memory files the agent wrote, and the worker sessions, which establish facts such as the language model that actually generated a candidate. We distinguish five stages of a revision: (i)~a diagnosis or proposal; (ii)~a defined component; (iii)~activation, after which the component belongs to the agent's composition; (iv)~actual use by subsequent search; and (v)~observed consequences. Only activated and used revisions are counted as changes to the agent. Several definitions never reached stage~(iii) or~(iv). In AHC027, for example, the agent noticed before activation that a new exploration branch was unreachable and replaced the definition; in AHC015 one version was superseded before any candidate was generated under it. Several revised loops also constructed worker instructions themselves rather than invoking the registered variation component, so the effective variation policy in those runs is read from the loop.

Scores are the values recorded during search: mean scores over the public cases for ALE-Bench, speedups measured during search for KernelBench, whose final results in Section~\ref{subsec:kernelbench_results} were evaluated through Modal, and the task objective for mathematical and systems tasks. They can differ from the private or independently re-measured results in Section~\ref{sec:experiments}. For each run we report $k_0$, the number of candidate attempts made under the initial composition before the first activated revision, and the share of the in-run improvement obtained afterwards,
\begin{equation}
\rho=\frac{f_T-f_{k_0}}{f_T-f_0},
\end{equation}
where $f_0$ is the starting program's score, $f_{k_0}$ the best score among the first $k_0$ candidates, and $f_T$ the final best score, all expressed in the direction in which the objective improves. Each run is a single trajectory in which revisions accumulate and interact, so temporal association between a revision and later progress does not identify its causal contribution. The analysis therefore characterizes how the agent developed its capabilities; it does not estimate the effect of individual revisions.

\subsection{Timing of Self-Modification and Search Progress}
\label{subsec:app_anatomy_timing}

Table~\ref{tab:app_anatomy_timing} reports, for every run, when self-modification began, how many revisions were activated, and how the best recorded score evolved. The agent first revised its composition after a median of four attempts and never later than the eighth. Revisions continued throughout the runs, with a median of 11 activations.

\begin{table}[!htbp]
\centering
\caption{Timing of self-modification and in-run progress. $k_0$: attempts under the initial composition; Rev.: activated revisions; $f_0$, $f_{k_0}$, $f_T$: starting, best-before-revision, and final best recorded scores; $\rho$: share of the in-run improvement obtained after the first revision. Arrows give the improvement direction. KernelBench speedups are those measured during search; the final results in Table~\ref{tab:kernelbench_results} were evaluated through Modal. PRISM recorded 16 candidates before its run ended; one circle-packing attempt was lost before evaluation.}
\label{tab:app_anatomy_timing}
\vspace{0.3em}
\resizebox{\linewidth}{!}{%
\begin{tabular}{@{}llrrrrrr@{}}
\toprule
\textbf{Task} & \textbf{Recorded objective} & $\mathbf{k_0}$ & \textbf{Rev.} & $\mathbf{f_0}$ & $\mathbf{f_{k_0}}$ & $\mathbf{f_T}$ & $\boldsymbol{\rho}$ \\
\midrule
AHC008 & Public score $\uparrow$ & 5 & 15 & 3{,}669{,}081 & 3{,}669{,}081 & 8{,}884{,}463 & 1.00 \\
AHC011 & Public score $\uparrow$ & 1 & 13 & 421{,}902 & 428{,}478 & 627{,}876 & 0.97 \\
AHC015 & Public score $\uparrow$ & 3 & 11 & 211{,}306 & 612{,}836 & 769{,}583 & 0.28 \\
AHC016 & Public score $\uparrow$ & 3 & 22 & 13{,}901{,}954 & 14{,}466{,}047 & 43{,}699{,}866 & 0.98 \\
AHC024 & Public score $\uparrow$ & 2 & 10 & 1{,}896.78 & 1{,}902.72 & 2{,}031.20 & 0.96 \\
AHC025 & Public cost $\downarrow$ & 3 & 12 & 3{,}806{,}444 & 3{,}806{,}444 & 636{,}087 & 1.00 \\
AHC026 & Public score $\uparrow$ & 7 & 15 & 9{,}258.80 & 9{,}303.34 & 9{,}488.82 & 0.81 \\
AHC027 & Public cost $\downarrow$ & 4 & 13 & 9{,}049{,}832 & 9{,}049{,}832 & 7{,}697{,}731 & 1.00 \\
AHC039 & Public score $\uparrow$ & 5 & 8 & 3{,}730.02 & 3{,}730.02 & 3{,}798.04 & 1.00 \\
AHC046 & Public score $\uparrow$ & 3 & 16 & 1{,}447.48 & 1{,}448.10 & 1{,}457.63 & 0.94 \\
\midrule
VGG16 & Speedup $\uparrow$ & 4 & 11 & $0.97\times$ & $1.58\times$ & $3.29\times$ & 0.74 \\
LayerNorm & Speedup $\uparrow$ & 8 & 11 & $0.99\times$ & $0.99\times$ & $19.35\times$ & 1.00 \\
ConvDiv & Speedup $\uparrow$ & 3 & 9 & $0.95\times$ & $0.95\times$ & $5.93\times$ & 1.00 \\
ConvMax & Speedup $\uparrow$ & 4 & 9 & $1.01\times$ & $1.31\times$ & $3.20\times$ & 0.86 \\
\midrule
Circle packing & Sum of radii $\uparrow$ & 8 & 9 & 0.9598 & 2.6153 & 2.6353 & 0.01 \\
Heilbronn triangle & Minimum area $\uparrow$ & 8 & 12 & 0 & 0.033559 & 0.036504 & 0.08 \\
Third-order autocorr. & $C_3$ $\downarrow$ & 4 & 10 & 2.5168 & 1.5977 & 1.4563 & 0.13 \\
\midrule
PRISM & Score $\uparrow$ & 6 & 2 & 21.892 & 26.256 & 26.256 & 0.00 \\
EPLB & Score $\uparrow$ & 2 & 6 & 0.12671 & 0.12671 & 0.14740 & 1.00 \\
LLM-SQL & Score $\uparrow$ & 3 & 11 & 0.67601 & 0.71107 & 0.73687 & 0.42 \\
Cloudcast & Transfer cost $\downarrow$ & 4 & 7 & 1{,}035.14 & 627.13 & 618.06 & 0.02 \\
\bottomrule
\end{tabular}%
}
\end{table}

Three regimes are visible. In seven runs (AHC008, AHC025, AHC027, AHC039, EPLB, LayerNorm, ConvDiv), no candidate produced by the initial composition improved on the starting program, so all progress followed self-modification. In these runs the initial failures were systematic rather than incidental: undirected changes that regressed (AHC008, AHC025, AHC027, AHC039), workers unaware of the output contract (EPLB), and kernels misrouted by the evaluation harness (LayerNorm, ConvDiv). In a second group, which includes most ALE-Bench and kernel runs, a few early attempts improved modestly and most of the gain followed the first revisions. In a third group (circle packing, Heilbronn, third-order autocorrelation, PRISM, Cloudcast, AHC015, LLM-SQL), the starting program was weak relative to designs reachable in a few attempts, and the first candidates closed most of the gap; for example, circle packing improved from a radius sum of $0.96$ to $2.60$ within three attempts. In this regime the later search was a long refinement near a strong configuration, and the agent's revisions concentrated on breaking plateaus. Several plateaus ended after specific revisions: Heilbronn held its record for 35 attempts until the agent began granting each extension a compute window derived from its parent's own budget; third-order autocorrelation held its record for 41 attempts until the agent supplied workers with the source of the strongest other candidates; AHC027 escaped a 20-attempt plateau through a recombination attempt after it spread selection away from a saturated leader; and circle packing advanced by less than 0.0001 over 17 attempts until the agent began supplying its best arrangement to workers as verified data. In Cloudcast, the record reached at attempt 24 was not exceeded; the agent's diagnostics showed that the dominant egress terms had converged across independent designs, and the remaining attempts were spent on bottleneck-targeted refinement and verification.

\subsection{Cross-Task Inventory}
\label{subsec:app_anatomy_inventory}

Table~\ref{tab:app_anatomy_matrix} records, for each run, which capability categories were changed by an activated revision that governed subsequent search. Every run revised candidate variation and maintained persistent memory, so these columns are omitted. Revisions that changed only population capacity are not counted as population management.

\begin{table}[!htbp]
\centering
\caption{Capability categories revised in each run ($\bullet$). Loop: restructured evolutionary loop or concurrency; Sel.: parent selection; Pop.: survival or archive policy; Rec.: recombination between candidates; Val.: validation or repair before a candidate is recorded; Rep.: repeated measurement used in ranking; Diag.: agent-built diagnostic or measurement tools; Model: worker language model; Wd.: explicit withdrawal or reversal of an earlier revision.}
\label{tab:app_anatomy_matrix}
\vspace{0.3em}
\small
\begin{tabular}{@{}lccccccccc@{}}
\toprule
\textbf{Task} & \textbf{Loop} & \textbf{Sel.} & \textbf{Pop.} & \textbf{Rec.} & \textbf{Val.} & \textbf{Rep.} & \textbf{Diag.} & \textbf{Model} & \textbf{Wd.} \\
\midrule
AHC008 & & $\bullet$ & & $\bullet$ & & & $\bullet$ & & \\
AHC011 & $\bullet$ & $\bullet$ & $\bullet$ & & $\bullet$ & $\bullet$ & $\bullet$ & & \\
AHC015 & $\bullet$ & $\bullet$ & $\bullet$ & $\bullet$ & & & $\bullet$ & & $\bullet$ \\
AHC016 & $\bullet$ & $\bullet$ & $\bullet$ & $\bullet$ & $\bullet$ & $\bullet$ & $\bullet$ & $\bullet$ & $\bullet$ \\
AHC024 & $\bullet$ & $\bullet$ & & $\bullet$ & & & & & \\
AHC025 & $\bullet$ & $\bullet$ & $\bullet$ & $\bullet$ & & $\bullet$ & $\bullet$ & & $\bullet$ \\
AHC026 & $\bullet$ & $\bullet$ & $\bullet$ & $\bullet$ & $\bullet$ & & $\bullet$ & & $\bullet$ \\
AHC027 & $\bullet$ & $\bullet$ & & $\bullet$ & $\bullet$ & & $\bullet$ & & $\bullet$ \\
AHC039 & $\bullet$ & $\bullet$ & & & $\bullet$ & & $\bullet$ & & \\
AHC046 & $\bullet$ & $\bullet$ & & $\bullet$ & & $\bullet$ & $\bullet$ & $\bullet$ & $\bullet$ \\
\midrule
VGG16 & $\bullet$ & $\bullet$ & $\bullet$ & & & & $\bullet$ & & $\bullet$ \\
LayerNorm & & $\bullet$ & & & & & $\bullet$ & & $\bullet$ \\
ConvDiv & $\bullet$ & $\bullet$ & $\bullet$ & $\bullet$ & & $\bullet$ & $\bullet$ & & \\
ConvMax & $\bullet$ & $\bullet$ & & $\bullet$ & & & $\bullet$ & & \\
\midrule
Circle packing & & & & $\bullet$ & $\bullet$ & & $\bullet$ & & $\bullet$ \\
Heilbronn triangle & & $\bullet$ & $\bullet$ & $\bullet$ & & & $\bullet$ & & $\bullet$ \\
Third-order autocorr. & $\bullet$ & $\bullet$ & & $\bullet$ & & & $\bullet$ & $\bullet$ & $\bullet$ \\
\midrule
PRISM & & $\bullet$ & $\bullet$ & & & & $\bullet$ & & \\
EPLB & $\bullet$ & & & & & & $\bullet$ & & \\
LLM-SQL & $\bullet$ & $\bullet$ & $\bullet$ & $\bullet$ & & & & & \\
Cloudcast & $\bullet$ & $\bullet$ & & & & & $\bullet$ & & $\bullet$ \\
\midrule
\textbf{Runs} & 16 & 19 & 10 & 14 & 6 & 5 & 19 & 3 & 12 \\
\bottomrule
\end{tabular}
\end{table}

The agent did not apply a fixed repertoire uniformly, and no two runs revised the same set of categories: AHC016 revised every category, whereas EPLB left selection and survival unchanged and concentrated on worker directions and loop concurrency, and LayerNorm left the loop unchanged while revising selection three times and worker guidance eight times. The categories that remained unchanged were those for which the agent's diagnoses did not identify a limitation.

\subsection{Candidate Variation}
\label{subsec:app_anatomy_variation}

\paragraph{Calibrating the scale of variation.}
The initial workers frequently replaced most of their parent. The first AHC025 children changed 313--1172 lines and all regressed; the first AHC024 children grew a 480-line program to 801 and 1163 lines, one of them scoring 1.0; none of the five initial AHC008 children improved on the starting score of 3.67 million, and four scored at most 0.89 million, two after near-total rewrites. The agent's first response was usually to request focused edits. It then calibrated this constraint from evidence rather than leaving it fixed. In AHC008, bounded edits kept all six children of the next interval within 0.55--1.35$\times$ of their parent's score, two of them setting a new best; the agent then permitted one larger change confined to a single subsystem after a 7-line edit had gained 35\%. Near the end of the run it measured improvement rates by edit size: edits of at most 20 lines improved their parent in 11 of 18 attempts (mean score ratio 1.013), edits of 20--140 lines in 24 of 61 (1.06), and edits of more than 140 lines in 1 of 11 (0.68). Having also observed that an earlier mandate for structural change at the frontier had produced an 88-line change that lost 8\% and a 72-line change that lost 54\%, it bounded the size of frontier changes accordingly. In LayerNorm, the opposite trend occurred: once candidates that rewrote their internal plan selection scored $17.09\times$ and $17.63\times$ while single changes matched the leader, the agent required exactly one change per candidate.

\paragraph{Task-specific operators.}
Beyond the generic distinction between refinement and redesign, the agent introduced operators tied to the structure of each task's evidence. In AHC011, per-case profiles showed that the starting program never completed a tree, so the objective could not enter its upper branch; directions then targeted the number of completed trees and later the number of moves. In AHC015, leading programs ran at 1.6\,s of a 2.0\,s limit, and \emph{robustify} attempts were assigned to the best candidate whose worst case exceeded 1.1\,s; one such child kept nearly the same score while reducing its worst case from 1.6\,s to 0.96\,s. In AHC046, \emph{economy} removed actions that a finished plan need not print, \emph{extend} asked for decisions the current plan could not express, and \emph{cheapen} sought value-preserving reductions in repeated work, after the agent observed that more alternatives could be evaluated per leg within the fixed time. EPLB introduced \emph{ablate}, which changes exactly one constant, branch, clamp, tie-break, or estimator. Heilbronn introduced an \emph{extend} contract that reproduces the parent's computation at its original scale and random-draw order, checkpoints the result, and only then adds work, and an \emph{audit} mode after a worker discovered that a swallowed exception had prevented one search move from ever running. VGG16 introduced an \emph{audit} role for the same reason, and the agent later gave it half of all attempts after eight audit candidates all scored at or above the frontier of their batch. AHC039 replaced undirected variation with assigned \emph{angles} on distinct aspects of the program (proposal moves, starting shape, time budget, objective exactness, and acceptance), later chosen by a softmax over measured parent-to-child changes, and finally required each change to be one that cannot lower the score after observing that every win had changed behavior only where it was demonstrably wrong.

\paragraph{Recombination.}
Table~\ref{tab:app_anatomy_recombination} summarizes the recombination mechanisms. They differ in what is transferred, which determines how much the worker must reconstruct. Transferring a complete donor program gives the most information but invites overwriting; the AHC024 instructions therefore evolved from a general merge request to a part-by-part comparison that retains the parent's version when alternatives cannot be judged. Transferring only what a donor added relative to its own parent (AHC027) isolates a mechanism; both of the first two such attempts transplanted the same exactly scored 2-opt move into a different branch and improved, and a later one ended a plateau by transplanting an exhaustive deletion scan. Transferring change lists (ConvDiv) relied on the agent's observation that candidates internally time their variants and keep the fastest, so a candidate that tries the union of two change lists should not be slower than either. Transferring structured lists of measured gains (AHC046) followed a worker's report that the shared notes misattributed which gains were present in the leading lineage. Transferring design reports (AHC025, AHC026, Heilbronn) carries a rationale without code; in AHC026 the agent replaced report-based synthesis with full-source transfer after two neutral outcomes. Donor choice was repeatedly revised to avoid recombining near-identical programs: AHC015 noted that merging near-identical descendants gained nothing, AHC024 moved from the single best donor to rank-sampled donors, AHC008 listed the strongest measured changes from branches outside the parent's ancestry so that compatible improvements could be grafted, and ConvMax paired parents with donors from a different implementation family after repairing its own family detector.

\begin{table}[!htbp]
\centering
\caption{Recombination mechanisms introduced by the agent.}
\label{tab:app_anatomy_recombination}
\vspace{0.3em}
\small
\begin{tabular}{@{}p{0.16\linewidth}p{0.16\linewidth}p{0.27\linewidth}p{0.31\linewidth}@{}}
\toprule
\textbf{Transferred content} & \textbf{Runs} & \textbf{Donor choice} & \textbf{Recorded observation} \\
\midrule
Complete donor source & AHC015, AHC016, AHC024, AHC026, circle packing, third-order autocorr., LLM-SQL, ConvMax & Best non-ancestor; best solution in the run; rank-sampled top donors; per-dataset champions; strongest candidate of another implementation family & AHC024 crossover children gained $+4.3$ and $+3.1$ points on average over two intervals; the AHC015 final lineage contains four hybridization steps; LLM-SQL's first recombinant became the ancestor of the entire frontier \\
Donor contribution relative to its own parent & AHC027 & Best live candidate that is neither the parent nor an ancestor & Both of the first two attempts improved ($+1.28\%$, $+0.48\%$); a later attempt ended a 20-attempt plateau ($+1.31\%$) \\
Change lists & ConvDiv & Three strongest other candidates & Merging non-conflicting increments while keeping the parent as base \\
Structured list of measured gains & AHC008, AHC046 & Measured gains from branches outside the parent's ancestry (AHC008) or absent from the leading lineage (AHC046) & Merge items assigned in rotation; leading lineage recovered gains made on side branches \\
Design reports & AHC025, AHC026, Heilbronn & Strong candidate outside the parent's ancestry; three best non-parent candidates & AHC026 report-based synthesis was replaced by full-source transfer after two neutral outcomes \\
\bottomrule
\end{tabular}
\end{table}

\paragraph{Validation and repair before recording.}
Workers could not execute code, so the agent supplied the checks they lacked. AHC011, AHC026, and AHC039 compiled candidates before official evaluation and spent one or two bounded repair attempts on failures; in AHC039 the agent traced two zero scores to a compiler that accepted constructs the judge's compiler rejected, and thereafter compiled with the judge's own compiler and flags. AHC026, AHC027, and AHC039 additionally ran candidates on a small set of cases and requested a repair when a candidate was illegal, crashed, or fell more than 5\% (AHC027) or 3\% (AHC039) below its parent. In AHC039 this became a \emph{completion} round in which the worker must locate the defect in its own change rather than start again; one candidate recovered from 3003 to 3600 on the check cases. AHC011 added a behavioral regression check triggered by a catastrophic first score, later raised to any score below 92\% of the parent estimate, and saved the pre-repair program so that a harmful repair could be rolled back; one repair that would have reduced a candidate's mean to zero was reverted. AHC016 repaired compilation, execution, and protocol failures in place before recording. AHC027 and AHC024 addressed candidates that were byte-identical to their parent, AHC027 by treating an unchanged file as a repair trigger. Circle packing checked each candidate against the evaluation contract and initially sent violations to a repair worker; after four of seven candidates spent that repair on cosmetic differences, it replaced the worker with deterministic restoration of the fixed region, and later changed the restoration to relocate, rather than delete, code that workers had placed outside the editable region.

\paragraph{Allocation from measured yield.}
The agent repeatedly computed per-operator outcomes and reallocated attempts. In AHC015 its own table recorded parent-relative improvement for 23 of 34 exploitation, 10 of 20 exploration, 23 of 32 hybridization, and 0 of 5 robustification attempts; hybridization rose from none to five of twelve planned attempts, restarts were removed after converging below the frontier, and exploration was briefly increased to four of twelve after a stall. VGG16 measured mean scores of 3.17 for audit, 3.16 for refinement, and 2.80 for redesign from compact ancestors, and removed the last role for the final attempts. LLM-SQL increased exploration and speed-oriented recombination after finding that balance and exploration children most often exceeded their parents. Third-order autocorrelation removed three exploration directions whose candidates reached $C_3$ values between 1.57 and 2.62 while exploitation stayed between 1.457 and 1.466. In EPLB, direction weights were $0.30+(w+1)/(n+2)$, where $w$ of $n$ children improved on their parent; the agent then diagnosed that this favored directions that beat weak parents early and changed the weight to $0.25+(w+1)/(n+2)+1.2\cdot\mathbb{1}[\text{best child}\ge f^\star-0.0004]$, where $f^\star$ is the run's best score. LayerNorm alternated adaptively between refinement and structural exploration, switching to exploration after three consecutive attempts failed to improve on the best.

\subsection{Parent Selection}
\label{subsec:app_anatomy_selection}

\paragraph{Sharpening selection pressure.}
Fitness-proportional sampling was replaced in most runs because recorded scores differed little relative to their magnitude. The replacements included tournaments of three or four, usually with a uniform share of 10--20\% (AHC027, AHC046, Heilbronn), rank weights $1/(r+1)$ (AHC024, PRISM), $1/r^2$ (AHC027), and $\max(0.05,0.45^{r})$ (third-order autocorrelation), merit-proportional sampling within the top half of the population (AHC015), and arms that return the current best a fixed fraction of the time (AHC026, AHC039). In AHC027 the agent recalculated the consequence of a larger population: a tournament of three selected the best candidate only about 27\% of the time, and rank weighting restored this to about 58\%. AHC011 replaced roulette selection with an explicit rank plan in which each group of exploitation attempts drew twice from each of the top two candidates and once from the third and fourth. Other schemes drew from an elite subset a fixed fraction of the time: AHC025 chose uniformly among the best three candidates with probability 0.75, AHC039 returned the best living candidate with probability 0.5 and later 0.6, and AHC016 alternated between the top two candidates and the whole active population before drawing from the top two on three of every four selections. Cloudcast retained fitness-proportional sampling but drew a uniformly random parent on every third selection.

\paragraph{Spreading selection when concentration fails.}
Concentration repeatedly produced redundant attempts, and the agent introduced mechanisms to spread them. In AHC025, a rolling set of six workers all started before any child was recorded, so a rule favoring the best unexpanded parent sent five of seven attempts to the same parent, none of which improved it. The agent moved selection into the loop so it could count children still in progress and weighted parents by $m/(1+0.7\,(n_{\text{recorded}}+n_{\text{in flight}}))$ over the top four, with 15\% of selections from the remaining population. In AHC027, rank weighting sent six of eight attempts to a saturated leader, all of which regressed, and the agent multiplied a candidate's weight by 0.3 for each of the last four selections that chose it. LayerNorm introduced a 1--1.75$\times$ weight for candidates structurally distant from the best, then replaced it with a 2.5$\times$ weight on the leader after observing that the novelty bonus spent attempts on sibling families while the leader was the only lineage still improving. AHC008 revised selection four times: it first demoted parents whose recent attempts had not improved them, then combined merit-based exploitation (75\%) with exploration of subtrees whose topics had not yet been tried (25\%), added a mode that selects outside the top three after observing that spreading parents within the frontier did not spread ideas, and finally made that mode conditional on the leader stalling once the next breakthrough came from the leader's own child. ConvDiv made broadening conditional in a similar way: it concentrated each batch on lines within 60\% of the best line and reopened the whole population once the best score had not changed for ten selections. Several loops also prevented redundant assignments directly: AHC024 hid parents that were already being expanded from the selector, ConvDiv added an anti-repeat rule so that a batch spreads across parents, LLM-SQL and EPLB redrew duplicate parents, and AHC011 avoided assigning an entire group of attempts to a single parent when another useful candidate was available.

\paragraph{Coupling parents with operations.}
Parent choice also became tied to the operation the parent would receive. The agent chose parents for specific operations: compact programs for redesign and the current best for small refinements (AHC026); zero-scoring candidates for repair (VGG16, and AHC026, whose repair arm fell from 15\% to 8\% of selections and was then removed after its only product scored 1113); the best candidate whose worst case exceeded 1.1\,s for \emph{robustify} (AHC015); and, in LLM-SQL, two attempts per wave on the leading candidate and the culled historical champion of the dataset on which the current best lagged most, used as the parent of an archive-rebreeding attempt with the current best as donor. Conversely, the agent assigned operations from a parent's standing or history. AHC015 assigned \emph{rethink} to a parent with at least three children of which none had improved and exploration to heavily expanded parents; AHC024 and AHC046 assigned repair to failed parents; Cloudcast assigned refinement to the leader and redesign to parents more than 35\% costlier; AHC016 ordered the selected parents so that its leading directions went to the strongest ones; and ConvMax assigned exploitation, cross-family recombination, redesign, and exploration by parent rank, so that every batch contained each mode. Selection thus came to depend not only on how good a candidate was, but on what the next attempt was meant to do with it.

\paragraph{Selection on attributes beyond fitness.}
The agent selected on quantities that its diagnostics associated with future progress. AHC026 reserved 30--55\% of selections for compact programs. ConvDiv scored a lineage by the best of its own score and its best child's score; after measuring that this kept a $1.13\times$ parent ahead of a $2.85\times$ candidate and produced the batch's two worst results, the agent reduced productivity to 5\% of the weight. AHC011, AHC016, AHC025, and ConvDiv ranked candidates by means over repeated official evaluations. ConvDiv also selected on its own diagnosis when it contradicted the official score: candidates recorded at zero only because of the harness's import rule kept a decaying priority based on the agent's own measurement of their speed. LLM-SQL assigned parents by walking its behavioral niches, one leading member per niche, before falling back to its selector. VGG16 and LayerNorm kept failed candidates selectable: VGG16 spent about 18\% of selections on children of failed candidates paired with a repair role, and LayerNorm assigned zero-scoring candidates a small floor because, as its diagnosis recorded, those candidates held the real designs. PRISM ordered candidates with identical scores by simplicity, and VGG16 at one stage reserved selections for compact ancestors, a policy it narrowed and then removed after thirteen such redesigns failed to reach their batch frontier. Several runs also selected the starting program again as a restart parent (AHC015, AHC016, ConvMax); two of these restart policies were later removed.

\subsection{Population Management}
\label{subsec:app_anatomy_population}

Ten runs replaced top-$k$ survival. Their criteria reveal what the agent judged worth preserving in each task:
\begin{itemize}
    \item \textbf{Behavior.} LLM-SQL grouped candidates by their vector of per-dataset hit rates and an execution-time band, retained the best candidate of each niche, and filled remaining places by fitness. Because near-duplicates were still counted as distinct, it coarsened the signature from three to two decimal places and widened the runtime band from two to five seconds.
    \item \textbf{Search direction.} AHC015 retained, in addition to the top $k$, the best candidate produced by each non-exploitation direction, provided it reached 55\% of the best score.
    \item \textbf{Ancestry.} Heilbronn retained the strongest candidate and then candidates whose three-generation ancestries differed.
    \item \textbf{Program size.} AHC026 retained the top $k-1$ candidates plus the strongest program below 27.5\,KB, reviving it if necessary.
    \item \textbf{Repairable failures.} VGG16 retained the two most recent failed candidates so that repair roles had a parent to start from.
    \item \textbf{Measured estimates.} AHC011, AHC016, and AHC025 ranked survivors by repeated-measurement estimates. AHC011 also retained candidates within 1.5\% of the cutoff after a candidate trailing by 0.1\% had been discarded, and AHC016 kept the best recorded candidate whenever it was still alive.
    \item \textbf{Simplicity.} PRISM ranked survivors with equal scores by the number of non-comment source lines.
    \item \textbf{Priority and archive.} ConvDiv ranked survivors by a blended priority and kept every candidate within 90\% of the champion selectable through an elitist archive.
\end{itemize}
Several runs treated the full history, rather than the living population, as the relevant pool. AHC025's corrected survival rule revived culled candidates whose repeated measurements placed them in the top set, and the final best descends from one of them. LLM-SQL observed that the only candidate holding the best hit rate on one dataset had been culled long before, although its single use as a parent had produced the ancestor of the entire frontier, and added an attempt each round that rebreeds the historical champion of the dataset on which the current best lags most.

\subsection{Evolutionary-Loop Restructuring}
\label{subsec:app_anatomy_loop}

The agent revised its evolutionary loop at two levels: the cadence at which it interrupted search to reassess itself, and the organization of search within each interval.

\paragraph{Cadence of self-assessment.}
Search proceeded in calls whose size the agent chose, and the agent revised itself between calls. Across the 21 tasks, it made a median of 13 search calls (range 2--18). Its first call was small, with a median of four attempts, and calls lengthened as search progressed: on 19 of 21 tasks, calls in the last third of the run were larger on average than those in the first third. VGG16 moved from calls of 4--9 attempts to calls of 18, EPLB from 2--4 to 15--16, and AHC024 from 2 to 14. Of the 244 calls after the first, 196 (80\%) followed at least one revision. The agent therefore reassessed itself frequently while diagnosing its initial composition and allowed longer intervals once a composition had proven productive, while still revising itself before most intervals.

\paragraph{Concurrency.}
On 16 tasks the agent restructured how attempts were scheduled within a call. The recorded diagnoses consistently identified worker generation as the bottleneck: a worker took about 3 minutes in EPLB, 6 in ConvDiv and ConvMax, 9 in LLM-SQL, and 10--13 in AHC024 and AHC039, against evaluations of seconds to about a minute. The resulting loops took several forms:
\begin{itemize}
    \item \textbf{Synchronized batches} of three to twelve concurrent workers followed by evaluation (AHC011, AHC016, AHC046, third-order autocorrelation, LLM-SQL, Cloudcast, ConvDiv, ConvMax), with widths revised during the run, for example from six to eight to twelve and back to eight in AHC016.
    \item \textbf{Pipelines} that keep workers active while earlier candidates are evaluated (AHC039, EPLB, VGG16), and \textbf{streaming refill}, in which a new worker starts as soon as any worker finishes (AHC015, AHC024, AHC025, VGG16). VGG16 progressed from batches to a deeper pipeline to streaming refill, each time after measuring idle evaluation time between batches.
    \item \textbf{Validated loops} that check and, if necessary, repair each candidate before evaluation (AHC026).
\end{itemize}
The agent measured the effect of these changes: VGG16 recorded a 2.4-fold increase in throughput after its first pipelined loop, and AHC024 completed eight attempts in about 8.3 minutes, when a single worker had taken 10--13 minutes. Official evaluations were serialized in most runs so that timing-sensitive scores were not measured under contention; AHC026 placed its own development checks in the same serialized queue for the same reason. AHC016, third-order autocorrelation, and LLM-SQL evaluated in parallel.

\paragraph{Stages within the loop.}
The agent also changed what happened between the generation of a candidate and its entry into the population. In AHC011, concurrent drafts passed through a compilation check with a bounded repair, then a behavioral check triggered by a weak first score, with the pre-repair program saved and restored if repair made it worse, before three and later five official samples were taken; between batches, the loop resampled leading candidates whose estimates rested on few samples. AHC026 compiled each candidate and ran a local rule and time check inside the serialized evaluation queue, followed by one repair attempt. AHC039 compiled with the judge's own compiler, replayed each candidate on ten public cases to check legality, and allowed up to two completion rounds before evaluation. AHC016 repaired compilation, runtime, and protocol failures in place and re-evaluated before recording, and AHC046 replaced single evaluation with three and later four repeated evaluations whose mean became the recorded score. Several loops also absorbed operations that the initial composition kept separate, choosing parents or constructing worker instructions within the loop itself (AHC011, AHC015, AHC025, AHC046, third-order autocorrelation). The loop thus became the place where checks, measurements, and repairs were composed into a single procedure for producing a recorded candidate.

\paragraph{Coordination and freshness of information.}
Concurrency introduced coordination problems that the agent diagnosed and repaired. Workers started from the population as it was recorded when they began, so concurrent attempts could not see one another: in AHC027, four of eight simultaneous attempts made the same change, and the agent bounded concurrency to three and constructed each worker's instructions only when it started, so that later workers received earlier results; in AHC046, all parents of a batch were chosen before any of its children was recorded. Selection that ignored work in progress concentrated attempts: AHC025 began counting children still being generated, and AHC024 hid parents that were already being expanded. Assignment rules also had to remain correct under concurrency: EPLB derived each worker's direction from the candidate's own index rather than from a counter shared by concurrent workers, and AHC046 corrected a rotation index that had skipped directions within a batch. Concurrency was accordingly treated as a quantity to be measured rather than maximized. AHC026 returned from eight to five workers after measuring fewer candidates per minute with eight, whereas AHC039 increased its workers from four to six, arguing that worker time rather than evaluation limited the attempts that the remaining budget could buy.

\paragraph{Robustness of the loop.}
As the loop grew more complex, the agent hardened it against its own failures. After a setup error in a new recombination component consumed an allocated attempt without producing a candidate (AHC016), and after an initialization-order error terminated a whole search call (circle packing), it made such failures non-fatal so that an allocated attempt would still be evaluated. AHC015 and LLM-SQL retried an evaluation once after an exception, and AHC024 stopped launching workers after three consecutive evaluation exceptions.

\paragraph{Schedules conditioned on the remaining attempts.}
Late in several runs, the agent conditioned its schedule on the attempts that remained. AHC027 disabled deep redesign once 14 or fewer attempts remained, on the grounds that a long shot needs follow-up attempts to pay off; VGG16 restricted its final nine attempts to audit and refinement; AHC026 and AHC015 narrowed their final portfolios to the operators with the most reliable recent yield.

\subsection{Worker Specialization and Model Allocation}
\label{subsec:app_anatomy_workers}

Table~\ref{tab:app_anatomy_roles} groups the worker roles recorded across runs by function. Roles were defined by the main agent and differed in their instructions, their parent assignment, and in some runs their donor material or language model.

\begin{table}[!htbp]
\centering
\caption{Worker roles introduced by the agent, grouped by function. Names are those recorded in the runs.}
\label{tab:app_anatomy_roles}
\vspace{0.3em}
\small
\begin{tabular}{@{}p{0.16\linewidth}p{0.29\linewidth}p{0.31\linewidth}p{0.14\linewidth}@{}}
\toprule
\textbf{Function} & \textbf{Recorded roles} & \textbf{Purpose} & \textbf{Runs} \\
\midrule
Local refinement & refine, exploit, refine-core, refine-weak, refine-tune, deepen, marginal, surgical, informed refiner & Improve one identified weakness while preserving working behavior & Nearly all \\
Structural exploration & explore, explore-design, restructure, rethink, deep rethink, diverge, fresh, rebuild, break the plateau & Replace a central mechanism, accepting an initial loss & Nearly all \\
Restart & restart, divergence restart, baseline restart & Start a different design family from the original program with accumulated knowledge & AHC015, AHC016, ConvMax \\
Recombination & crossover, combine, hybridize, synthesize-design, bridge, merge, consolidate, combine-speed, archive-combine & Integrate mechanisms from another candidate & 14 runs \\
Repair and verification & repair, completion, repair-with-donor, audit, verify configuration & Correct defects, find inert mechanisms, or test optimality claims & AHC011, AHC016, AHC039, VGG16, Heilbronn, LLM-SQL, Cloudcast \\
Objective-targeted & completion and move count, economy, cheapen, extend, robustify, balance, instance gap, bottleneck, ablate, simplify & Act on a component of the objective identified by diagnostics & AHC011, AHC015, AHC046, Heilbronn, LLM-SQL, Cloudcast, EPLB, PRISM, third-order autocorr. \\
\bottomrule
\end{tabular}
\end{table}

\paragraph{Personas and family awareness.}
AHC015 distinguished a research persona for exploration attempts, which expects a structurally different design and tolerates an initial loss, from the engineering persona used for other attempts. ConvMax tagged candidates by implementation family, reported per-family best scores and failure counts to workers, and assigned modes by parent rank so that each batch contained refinement, cross-family recombination, redesign, and exploration.

\paragraph{Model allocation.}
Worker records confirm three runs in which the agent changed the generating model. AHC016 assigned \promodel{}, which its model catalog described as suited to difficult reasoning, to selected roles for ten candidates; the agent then measured that these candidates fell in the same score band as those of the default model while a batch took 45 rather than 20 minutes, and returned all workers to the default model. Third-order autocorrelation routed bottleneck and plateau-breaking roles to \promodel{} for 25 of its 100 workers and kept this routing to the end; its final best was produced by such a worker from a parent generated by the default model. AHC046 routed extension and merging roles to \promodel{} for three candidates, one of which set the final record and one of which failed to compile, and returned to the default model for its last four attempts.

\subsection{Memory}
\label{subsec:app_anatomy_memory}

On every task the agent maintained memory that later workers received, but it built several distinct kinds of memory and revised each of them during search. Table~\ref{tab:app_anatomy_memory} distinguishes them by content and use. Their development shows memory changing from a passive store of past attempts into a signal that governed selection, assignment, and recombination.

\begin{table}[!htbp]
\centering
\caption{Kinds of memory developed by the agent.}
\label{tab:app_anatomy_memory}
\vspace{0.3em}
\small
\begin{tabular}{@{}p{0.15\linewidth}p{0.27\linewidth}p{0.22\linewidth}p{0.26\linewidth}@{}}
\toprule
\textbf{Kind} & \textbf{Content} & \textbf{Use} & \textbf{Tasks} \\
\midrule
Curated lessons & Findings distilled and repeatedly rewritten by the agent & Supplied to every worker & AHC011, AHC016, AHC024, AHC025, AHC026, AHC039, AHC046, VGG16, ConvDiv, ConvMax, EPLB, Cloudcast \\
Attempt reports & Each worker's account of its change, stored per candidate & Retrieved for later workers, often with the measured score & Most tasks \\
Lineage history & The parent's report, the change that created it, and the edits already tried from it & Supplied to the parent's children & AHC008, AHC024, AHC025, AHC027, AHC039, third-order autocorr., ConvMax \\
Outcome digests & Recent attempts with their change relative to their parent & Shows which kinds of change have paid & AHC015, AHC027, PRISM, EPLB, LLM-SQL, circle packing, ConvMax \\
Elite knowledge & Reasoning or source of the strongest candidates outside the parent's line & Transfers mechanisms across lineages & AHC015, Heilbronn, third-order autocorr., LayerNorm, circle packing \\
Structured knowledge & Topic labels, measured gains, implementation families, component-wise bests, per-dataset leaders & Read by selection, assignment, and recombination & AHC008, AHC046, ConvMax, Cloudcast, LLM-SQL \\
Measurement memory & Repeated scores and per-case diagnostic profiles & Ranking candidates; parent-specific briefs & AHC011, AHC016, AHC025, AHC039, ConvDiv, Cloudcast \\
Solution memory & The best solution itself, verified independently of the program that produced it & Floor preserved by later attempts & Circle packing \\
Search-behavior memory & Histories of allocation, assigned directions, timing, and operator yield & Diagnosing the agent's own components & AHC008, AHC015, AHC024, AHC026, EPLB, VGG16 \\
\bottomrule
\end{tabular}
\end{table}

\paragraph{Deciding what to store.}
The agent repeatedly revised what a memory entry should contain. LLM-SQL switched from storing the last 2{,}000 characters of each worker response to the first 6{,}000 after observing that a truncated ending carried less information than a concise opening summary. AHC015 extracted the section following each report's final marker, doubled the stored length, and paired every report with its measured score, so that later workers could separate realized gains from plausible but unsuccessful ideas. Heilbronn standardized the headings of worker reports so that its briefings could extract what the parent did, what was changed, and why the change was expected to help. The agent also wrote memory itself rather than only collecting it: before its first revision took effect, AHC015 recorded notes on the three initial candidates from its own reading of their code and outcomes, and curated lessons were rewritten rather than appended. In AHC011, eight successive versions tracked the number of completed trees, the move cost of competing completion strategies, and the regions in which failures concentrated.

\paragraph{Retrieval and its repair.}
Storing information did not guarantee that it reached workers, and the agent detected several retrieval failures from their effects. AHC027's digest took the first line of each report, which was always the same verification preamble, so its memory carried no content until the agent switched to the line describing the change. Third-order autocorrelation passed a candidate identifier in place of the parent's report. In AHC008, workers' topic labels were written in a form that the extraction missed, which left novelty-based selection operating on empty data until the agent noticed the absence of topics in its own selection log. In each case, the correction changed the information that later workers or components actually received.

\paragraph{Widening the scope of memory.}
The scope over which memory was consulted grew with the evidence of duplicated work. AHC008 first supplied each worker with the edits already tried on its parent; after five of ten attempts repeated one idea within a single lineage, it extended this list to the parent's whole ancestry, and after four cousins pursued the same idea under different parents, it added run-wide repetition detection. Third-order autocorrelation replaced a static lessons block with a lineage context of parent, grandparent, siblings, and children, and later added the source and rationale of the strongest candidates in other lines. LayerNorm embedded the leading candidate's full report in every worker's instructions so that, in the words of its diagnosis, the deepest reasoning in the run would propagate rather than be rediscovered. AHC046 replaced a single list of gains believed to be present in the leading lineage with a structured list after a worker found that a listed gain was absent from its parent, which allowed gains made on side branches to be merged into the leading line.

\paragraph{Memory as a control signal.}
Memory increasingly governed decisions rather than only informing workers. In AHC008, topic labels determined which subtrees the exploration mode treated as untried; in AHC046, each merge worker was assigned a specific entry from the list of measured gains; in ConvMax, implementation-family tags determined the donor of each cross-family recombination, once the agent had corrected a detector that placed every candidate in the same family; in EPLB, the stored outcomes of each direction set the weights of the direction bandit; and in AHC011, AHC016, and AHC025, stored repeated measurements determined parent selection and survival. The clearest case is circle packing. After diagnosing that the search stored only programs, whose time-dependent constructors did not reliably reproduce their results, the agent stored the best arrangement itself, verified its feasibility, and supplied its 26 centres and radii to every subsequent worker as a floor to preserve, re-basing it whenever a better arrangement was captured. Memory thereby changed from a description of past attempts into a starting point for future ones, and the 17-attempt plateau ended in the next interval.

\paragraph{Memory of the agent's own behavior.}
The agent also retained records of how its own components behaved, and several revisions originated there. Counting the directions actually assigned in AHC024 revealed that 28 of 32 attempts had received repair. AHC015 tracked outcomes by direction and removed restarts after all five restart attempts converged about 5\% below the frontier. VGG16 added a timestamped log of each phase of its loop and used it to identify idle evaluation time while whole batches of workers were still running, which motivated streaming refill. AHC026 added worker timing logs after its throughput measurement so that later concurrency decisions could be measured rather than estimated. This form of memory underlies the self-corrections documented in Section~\ref{subsec:app_anatomy_correction}.

\subsection{Measurement and Diagnostic Capabilities}
\label{subsec:app_anatomy_measurement}

\paragraph{Instruments built by the agent.}
The agent built tools that it then used to decide what to revise. They included per-case profilers (AHC011, AHC015, AHC016, AHC039, AHC046); development scorers validated against official scores, exactly in AHC008, within about 0.36\% in AHC027, and within about 20 points of 9{,}259 in AHC026; independent legality checkers (AHC039); a loss decomposition against an offline reference partition (AHC025); bounds and feasibility probes (PRISM, EPLB, Cloudcast); generators of held-out instances to check generalization (PRISM with five additional seeds, Cloudcast with 21 additional configurations); a verifier that extracted the best arrangement, checked its feasibility, and showed that its radii were already maximal for its centres (circle packing); per-configuration cost and bottleneck analyzers (Cloudcast); measured memory bandwidth (LayerNorm); and per-kernel time breakdowns of the leading candidate (ConvMax). The ConvMax breakdown attributed about 71\% of kernel time to the library convolution, 16\% to a layout conversion, and 12\% to the fused epilogue, with about 0.43\,ms per call outside kernels, and each worker was then asked to choose its single change from that breakdown.

\paragraph{Discovering the evaluation context.}
The agent analyzed how candidates were actually evaluated. In LayerNorm, ConvDiv, and VGG16, it found that the KernelBench harness detects Triton kernels with a rule that matches only an unindented import at the start of a line; guarded or indented imports silently route a candidate through an execution path in which its kernels fail. Seven of the first eight LayerNorm candidates failed for this reason alone, and one of them scored $12.44\times$ once the import was moved. In ConvDiv all three initial candidates were recorded at zero for the same reason, although the agent's own evaluation through the correct path found them correct, at $1.08\times$, $1.50\times$, and $0.98\times$. The agent recorded these facts in worker instructions. In AHC011 it found that the objective has a separate branch for complete trees that the starting program never reached. In EPLB it established that simulation time dominated the measured execution time, so balancedness gains outweighed algorithmic speedups.

\paragraph{Noise and repeated measurement.}
Where repeated evaluations of unchanged programs varied, the agent quantified the variation and adapted its decisions. ConvDiv re-evaluated a candidate recorded at $3.29\times$ and obtained 3.02, 3.02, 3.22, and 3.15, concluded that a single run is a poor ranking signal within a $\pm5\%$ spread, and ranked by repeated means. AHC025 observed that a candidate recorded at a cost of 1.16 million measured between 1.64 and 1.71 million on re-evaluation and introduced repeated sampling; AHC016 and AHC046 adopted three or four evaluations per candidate. Circle packing found that re-running its best program, whose constructor depended on elapsed time, returned 2.627558 and 2.628678 against a recorded 2.628570; instead of relying on the program to reproduce its result, it stored the arrangement itself as verified data for later attempts. Several runs ended by explicitly separating statistically indistinguishable leaders: in AHC011, interleaved re-measurement showed that differences among the leading candidates were smaller than their standard errors, and in AHC016 the best recorded candidate and the best estimated candidate differed.

\paragraph{Diagnosis of saturation.}
When diagnostics indicated that an objective was near its attainable value, the agent redirected search rather than continuing to refine. PRISM found that four independently constructed programs produced identical per-instance values on the 50 evaluation instances and on 250 instances from five other seeds, and a feasibility search reproduced the achieved value to within $2\times10^{-9}$ relative on the instances it checked; it concluded that the visible objective was saturated and redirected workers toward programs that match the score with fewer components and robust behavior on unseen input shapes, ranking equal-scoring candidates by simplicity. LayerNorm measured that its leading design already ran within about 4\% of the memory bandwidth it had measured as attainable and instructed workers that further tuning of that design could not help; it later identified, from the leading worker's report, serving the second read of the input from cache, against a 512\,MiB traffic floor, as the remaining lever. EPLB found that its top ten candidates shared identical balancedness and differed only in timing.

\begin{table}[!htbp]
\centering
\caption{Representative diagnoses recorded by the agent when defining a revision (verbatim excerpts).}
\label{tab:app_anatomy_diagnoses}
\vspace{0.3em}
\small
\begin{tabular}{@{}p{0.12\linewidth}p{0.15\linewidth}p{0.66\linewidth}@{}}
\toprule
\textbf{Run} & \textbf{Revision} & \textbf{Recorded diagnosis} \\
\midrule
AHC027 & Redirect redesign & ``Every attempt that changed how the route is CONSTRUCTED \ldots made the score worse (6 clear regressions, one -0.13\% wash), with a single exception (c000040, +0.93\%). Meanwhile almost every attempt that changed how the search IMPROVES the route \ldots helped.'' \\
AHC026 & Size-aware selection & ``every large redesign gain in this run started from a parent of at most \textasciitilde27 KB of source \ldots while all ten redesigns that started from a program of 28 KB or more produced no gain and several heavy losses.'' \\
AHC024 & Direction allocation & ``28 of 32 attempts received `repair' and only one `explore' \ldots so the intended mix never ran.'' \\
AHC025 & Survival & ``each cull kept the five WEAKEST alive candidates and dropped the strongest.'' \\
AHC027 & Bounded concurrency & ``four of its eight attempts \ldots independently made the SAME change \ldots and all four lost.'' \\
AHC039 & Variation direction & ``Every win in this run \ldots came from a change that cannot make the score worse.'' \\
AHC016 & Model allocation & ``the pro attempts landed in exactly the same band (36.2-36.9M estimated) but the batch took 45 minutes against 20 for an all-default batch \ldots The stronger model buys no quality here.'' \\
LLM-SQL & Archive rebreeding & ``the only candidate holding the run's best beer value (c000000, 0.707) was culled long ago \ldots yet its single appearance as a parent produced c000010, the ancestor of the whole current frontier.'' \\
LayerNorm & Structural novelty & ``three separate candidates scored exactly 18.95x, meaning their independent auto-tuners all selected the same already-proven plan. The run is no longer exploring designs, only which plan each candidate measures.'' \\
ConvDiv & Ranking & ``re-running the unchanged official evaluator on the same candidate file gives 3.02/3.02/3.22/3.15 for the candidate recorded at 3.29 \ldots a single official run is a poor ranking signal.'' \\
Heilbronn & Budget reader & ``the parent-budget reader parsed the report text and read 50s for c000012 (whose real budget is 205s).'' \\
ConvMax & Family detection & ``Family detection counted any library conv call, including one hidden in a never-taken fallback, so every candidate looked like the same family and the cross-family crossover never fired.'' \\
\bottomrule
\end{tabular}
\end{table}

\subsection{Self-Correction, Withdrawal, and Recursive Revision}
\label{subsec:app_anatomy_correction}

Every run contains at least one correction by the agent of its own instruments or revisions. Table~\ref{tab:app_anatomy_corrections} lists representative cases. Three features stand out.

First, the agent detected defects by comparing what its components actually did with what they were meant to do, using the records those components produced. The AHC024 direction history, the AHC026 parentage of the launchpad, the AHC027 memory digest, the AHC025 survival flags, and the AHC008 selection log each exposed a discrepancy that no score alone would reveal.

Second, the instruments used to decide further revisions were themselves revised, and the corrected instruments shaped later revisions. In AHC011, the regression check never fired because it counted legal outputs as completed trees; once corrected, it detected a child that had lost twelve completed trees relative to its parent, and its directions were updated with the corrected profile. In Heilbronn, the agent's estimate of each parent's compute budget determined the windows granted to extensions; correcting it twice preceded two consecutive records. In Cloudcast, an extended bottleneck analyzer identified the link that bounds transfer time, which became the next revision's direction. In AHC016, the estimator that informed selection had counted cached measurements as new samples until the agent required fresh measurements.

Third, withdrawal was a frequent outcome. Twelve runs explicitly removed, deactivated, or reverted an earlier revision. These included the circle-packing repair worker, replaced by deterministic restoration, the alternative worker model in AHC016 and AHC046, restart policies in AHC015 and AHC016, exploration directions in third-order autocorrelation, VGG16's compact-ancestor redesign, LayerNorm's novelty bonus, AHC026's repair arm and additional workers, and Cloudcast's verification-only portfolio. In Cloudcast, the agent reactivated an earlier version of its own component, so the population and diagnostics accumulated under the withdrawn version were carried into search under the restored one.

\begin{table}[!htbp]
\centering
\caption{Representative defects in the agent's own capabilities and their correction.}
\label{tab:app_anatomy_corrections}
\vspace{0.3em}
\small
\begin{tabular}{@{}p{0.12\linewidth}p{0.28\linewidth}p{0.25\linewidth}p{0.25\linewidth}@{}}
\toprule
\textbf{Run} & \textbf{Defect or limitation} & \textbf{How it was detected} & \textbf{Revision} \\
\midrule
AHC008 & Workers' topic labels were written in a form the parser missed, so novelty-based selection operated on empty data & Its own selection log showed no topics for any candidate & Tolerant label extraction; label requested first \\
AHC008 & Mandate for structural change at the frontier produced large losing changes & Improvement rates measured by edit size & Size of frontier changes bounded \\
AHC011 & Regression check counted legal but incomplete outputs as completed trees & Check never fired & Completion defined by the objective's branch \\
AHC011 & Repair could damage a working program & A repair reduced a candidate's mean to zero & Pre-repair program saved and restored \\
AHC016 & Initialization error in a new recombination component consumed an attempt without a record & Missing population record & Error corrected; setup failures no longer discard attempts \\
AHC024 & Repair override triggered for nearly every parent & Counting realized directions (28 of 32) & Threshold relative to the best; new divergence direction \\
AHC024 & All crossovers used the same donor & Donor audit & Rank-sampled donors among the top three \\
AHC025 & Survival kept the five weakest candidates & Strong candidates culled while a weak one survived & Correct ordering, repeated confirmation, and revival \\
AHC026 & Compact-parent arm never fired with a single compact candidate & Launchpad produced no children & Condition corrected; six children followed \\
AHC027 & Memory stored boilerplate first lines & Reading its own digest & Memory records the described change \\
AHC039 & Local compiler accepted code the judge rejected & Two official compilation errors & Checks use the judge's compiler and flags \\
Heilbronn & Parent budget read as 50\,s instead of 205\,s, later as an unrelated 999\,s constant & Comparison with the parent's source & Reader based on the parent's source with bounds \\
LLM-SQL & Repair aimed at the dataset with the lowest absolute rate, already near its ceiling & Per-dataset gaps & Repair targets the largest recoverable gap \\
ConvDiv & Productivity weighting favored an unproductive $1.13\times$ parent over a $2.85\times$ candidate & Two worst results of a batch & Productivity reduced to 5\% of the weight \\
ConvMax & Wrong tolerance in worker instructions; family detector misclassified all candidates & Failure diagnostics; cross-family pairing never occurred & Tolerance corrected; detector based on the candidate's own kernels \\
Circle packing & Contract enforcement deleted worker code outside the editable region, leaving a candidate without its entry point & A zero score; 14 of 44 candidates had triggered the enforcement & Code relocated into the editable region instead of deleted \\
Circle packing & Initialization-order error in a new version terminated a search call and lost an allocated attempt & The interrupted call & Constants reordered; enforcement made non-fatal \\
\bottomrule
\end{tabular}
\end{table}

\subsection{Representative Trajectories}
\label{subsec:app_anatomy_trajectories}

\paragraph{AHC008: memory scope and the measured economics of edit size.}
None of the five candidates produced by the initial composition improved on the starting score of 3.67 million; four scored at most 0.89 million, two after near-total rewrites. The agent required bounded edits that preserve the parent's working structure, and the next six children all stayed within 0.55--1.35$\times$ of their parent while two set new records. When two lineages independently made the same edit, it gave each worker the parent's report and the list of edits already tried on that parent. After 19 candidates, observing that selection placed about 90\% of its weight on the single best candidate while several parents had absorbed three or more fruitless attempts, it began demoting such parents. When eight near-identical candidates around 5.2 million then absorbed six of eight attempts, while the one attempt outside them produced the largest gain of the run, it split selection into a merit-based exploitation mode and a mode that explores subtrees with untried topics. It then built a development scorer that reproduced each candidate's official score exactly in about eight seconds, and used it to give workers their parent's mean score, its weakest cases, and the mechanism of the leading line. Its own selection log revealed that the exploration mode had been operating without topics because the labels were written in a form its parser missed, and it corrected the extraction. As duplicated ideas reappeared at wider scopes, first two generations up and then among cousins, it extended the memory from the parent to the whole ancestry and finally to the run, and it listed measured improvements from rival branches as candidates for grafting. An analysis by idea family showed that the three largest families were saturating while a mechanism tried once had gained 9\%, so the leading candidate received a mandate for structural change; when that mandate produced large losing edits, the agent measured improvement rates by edit size and bounded the frontier change. The record advanced from 3.67 to 4.96, 5.33, 6.66, 7.03, 7.11, 7.54, 8.13, 8.16, and finally 8.88 million, $2.42\times$ the starting score, with 38 of 100 candidates improving on their parent. The last 23 attempts produced no new record, an outcome the agent itself attributed to the structural mandate.

\paragraph{Circle packing: remembering the solution as data.}
The first eight attempts raised the radius sum from 0.96 to 2.615. The agent then observed that workers imposed a 20-second budget on themselves although the evaluator allowed about 360 seconds, and it gave them an explicit compute budget and timeout rule. When four candidates exceeded the time limit and one broke the fixed scaffolding, it stated the exact evaluation contract, added a digest of recent outcomes, and checked each candidate against the contract with one repair attempt. After four of seven candidates spent that repair on cosmetic differences, it replaced the repair worker with deterministic restoration of the fixed region and began giving every worker the best solution produced anywhere in the run together with recent findings. Its next version terminated a search call through an initialization-order error and lost one allocated attempt, and the agent made the enforcement non-fatal. As eight attempts clustered within 0.0003 of each other, it introduced exploitation, exploration, and recombination directions, withholding the best solution from exploration attempts so that they would not anchor on it. The decisive diagnosis followed: the leading arrangement was tight, with radii already maximal for its centres, the constructor depended on elapsed time, so re-running the best program gave 2.627558 and 2.628678 against a recorded 2.628570, and the search stored only programs, so every attempt re-derived the frontier. The agent extracted the best arrangement, verified it, and supplied the 26 centres and radii to workers as a guaranteed floor. After 17 attempts with almost no progress, the next interval recorded 2.635024, although the agent noted that its own re-run of that program scored lower. Observing that the breakthrough attempt had changed which circles touch, while twelve attempts that only adjusted positions returned the floor unchanged, it made structural change the leading direction and re-based the floor whenever a better arrangement was captured. It then found that its own contract enforcement, which 14 of 44 candidates had triggered, deleted code that workers had placed outside the editable region, leaving one candidate without its entry point, and changed the enforcement to relocate such code. With the captured floor rising from 2.6287 to 2.6352, records followed at 2.635093, 2.635175, 2.635252, and finally 2.635314.

\paragraph{AHC026: inventing a population rule from outcome data.}
The first seven attempts under the initial composition consisted of large rewrites, and one produced invalid output. The agent introduced tournament selection with a small arm for repairing failed candidates, role-specific workers separating refinement from redesign, and a compilation check that initially compiled the delimiters of the evolvable region and flagged every candidate until the agent corrected it. It then built a local simulator validated against official scores and used it in a validated loop that checks each candidate before evaluation; a candidate repaired by this check lies on the winning lineage. After measuring that eight workers produced fewer candidates per minute than five, it reverted. Having observed that two strong but unrelated designs could not exchange ideas, it added synthesis from the other line's design report and later from its full source. After about fifty attempts it related each redesign's outcome to its parent's source size, found that all large gains came from parents of at most about 27\,KB, and reorganized selection and role assignment around compact programs. Three further defects followed and were corrected in turn: survival had culled every compact candidate, the first launchpad was simply the smallest rather than the strongest compact program, and an off-by-one condition prevented the compact arm from firing. The final record ($9{,}488.82$ against a starting score of $9{,}258.80$) came from refinement on the main line, which also corresponded to the highest private performance among the compared methods on this task.

\paragraph{AHC027: memory, concurrency, and recombination.}
All four initial candidates were worse than the starting program, and three re-evaluations of the starting program established that the regressions exceeded evaluation noise. The agent introduced a memory digest of recent attempts with their parent-relative changes, a development check on eight public cases with one repair attempt for candidates more than 5\% worse than their parent, and tournament selection. It then found that its memory had stored only boilerplate, repaired it, and used the corrected memory to classify all attempts by the part of the solution they changed. This revealed that changes to route construction almost always regressed while changes to route improvement almost always helped, and it redirected both refinement and redesign accordingly; the next interval produced an exactly scored 2-opt move. It introduced recombination of a donor's contribution, which transplanted that move into a different branch. When four of eight concurrent workers made the same change, it bounded concurrency and constructed instructions only when a worker started. When rank weighting concentrated six of eight attempts on a saturated leader, it added a recency penalty, and the next interval escaped the plateau through recombination. Its final revision changed the portfolio once 14 or fewer attempts remained. The lineage of the final best passes through three candidates that were worse than their parents when produced and two recombination steps, and the run reduced recorded cost by 14.9\%.

\paragraph{LLM-SQL: behavioral niches and opportunity-based roles.}
The scalar objective combined five per-dataset hit rates with an execution-time term, and the agent's first revision exposed each parent's per-dataset profile, its weakest dataset, and the best rate observed on each dataset. Recombination drew donors that held the best rate on the parent's largest-gap dataset, and its first recombinant became the ancestor of the entire frontier. When several candidates repeated the same behavioral profile, the agent introduced niche-based survival over hit-rate vectors and execution-time bands, coarsened the niches after near-duplicates persisted, and walked the niches when assigning parents. It then compared the remaining accuracy and speed opportunities to choose roles, corrected repair to target the largest recoverable gap rather than the lowest absolute rate, and finally added archive rebreeding after recognizing that the starting program, culled long before, still held the best rate on one dataset. The recorded score rose from 0.676 to 0.737. One heuristic in this sequence, the comparison of speed opportunity, contained a sign error that the agent did not detect, so its elite attempts favored repair and exploration over speed-oriented recombination; explicit speed-recombination attempts remained in the portfolio.

\paragraph{Heilbronn: compute allocation, inheritance, and a self-corrected estimator.}
The agent observed that candidates used between 3 and 15 seconds of a 360-second allowance and introduced explicit computation windows. Escalating these windows produced no record and brought evaluations close to the time limit, so it reduced them and introduced briefings that carry the successful mechanisms of the three best non-parent candidates. While the record held for 35 attempts, it introduced an inheritance contract under which extensions first reproduce the parent's computation exactly and then add work, and subsequently derived each extension's window from the parent's own budget; the record improved in the first interval under the latter. A worker's report that a swallowed exception had disabled a search move prompted an audit mode. The agent then found that its budget reader had read 50 seconds for a parent whose actual budget was 205 seconds, corrected it, later corrected it again when it read an unrelated constant, and each correction preceded a new record. The final directions targeted inert operators and search neighborhoods whose step sizes were coarser than necessary. The recorded minimum triangle area reached 0.0365035.

\paragraph{LayerNorm and ConvDiv: from harness contracts to measured limits.}
In both kernel runs the initial composition produced no valid improvement. The agent traced the failures to the harness's import-detection rule, gave workers the exact evaluation contract and each parent's failure excerpt, and kept zero-scoring candidates selectable because they held the real designs. LayerNorm then produced a $15.98\times$ kernel in the next interval. When candidates plateaued, the agent measured the attainable memory bandwidth, showed that the leading design was already close to it, and propagated the leading worker's report to all workers. When three candidates scored exactly $18.95\times$, it diagnosed that their internal tuners selected the same plan and required each candidate to contribute a structure absent from the parent. It later found that internal tuners never timed late plans before their deadline and asked workers to measure their chosen structure first, and finally restricted workers to single changes after rewrites of the tuner regressed. ConvDiv progressed from $1.13\times$ to $3.29\times$ through harness-aware instructions, detection of candidates that silently executed their eager fallback, lineage-aware selection that it corrected after measuring its harm, and recombination of change lists. After measuring a $\pm5\%$ spread between identical evaluations, it ranked candidates by repeated means; the remaining 42 attempts raised the recorded speedup to $5.93\times$.

\paragraph{PRISM and Cloudcast: reasoning about attainable improvement.}
In both runs early attempts reached a strong configuration, and the agent's subsequent revisions addressed what further search could achieve. In PRISM, after a first revision introduced rank-based selection and workers informed by the exact pressure formula, four of six children tied the best score exactly. The agent built a per-instance analyzer with a placement-independent lower bound, a generator of instances from five additional seeds, and a feasibility probe; it questioned the probe's first result, which declared every instance improvable at a margin comparable to its numerical tolerance, and refined it by bisection before concluding that the objective was saturated. It then redirected selection, survival, and worker objectives toward simpler programs that preserve the score and behave safely on unseen inputs. In Cloudcast, cost fell from 1035.1 to 618.1 within the first 24 attempts. The agent then built per-configuration cost analysis, 21 held-out configurations, bounds, and a bottleneck analyzer, targeted the instance-cost term and the specific link bounding transfer time, tried a portfolio devoted entirely to verifying each configuration's optimality, and reverted to its previous composition when twelve verification attempts all tied the best.

\subsection{Scope of Interpretation}
\label{subsec:app_anatomy_scope}

The analysis covers one run per task, and revisions within a run were not applied in isolation. Improvements that followed a revision are therefore temporal associations; several were preceded by multiple simultaneous changes, and many recorded scores are subject to the measurement variation that the agent itself quantified. Some revisions contained errors the agent did not detect, such as the LLM-SQL sign error and arithmetic inconsistencies in some Cloudcast worker instructions, and some were never activated or never used. These limitations do not affect the central observations: the agent repeatedly revised its own search capabilities from measured evidence, monitored the behavior of the revised agent, and corrected or withdrew its revisions, including the diagnostic procedures through which it judged them.

%% file: references.bib
@misc{novikov2025alphaevolve,
      title={AlphaEvolve: A coding agent for scientific and algorithmic discovery}, 
      author={Alexander Novikov and Ngân Vũ and Marvin Eisenberger and Emilien Dupont and Po-Sen Huang and Adam Zsolt Wagner and Sergey Shirobokov and Borislav Kozlovskii and Francisco J. R. Ruiz and Abbas Mehrabian and M. Pawan Kumar and Abigail See and Swarat Chaudhuri and George Holland and Alex Davies and Sebastian Nowozin and Pushmeet Kohli and Matej Balog},
      year={2025},
      eprint={2506.13131},
      archivePrefix={arXiv},
      primaryClass={cs.AI},
      url={https://arxiv.org/abs/2506.13131}, 
}

@article{romeraparedes2024funsearch,
  title={Mathematical discoveries from program search with large language models},
  volume={625},
  ISSN={1476-4687},
  url={http://dx.doi.org/10.1038/s41586-023-06924-6},
  DOI={10.1038/s41586-023-06924-6},
  number={7995},
  journal={Nature},
  publisher={Springer Science and Business Media LLC},
  author={Romera-Paredes, Bernardino and Barekatain, Mohammadamin and Novikov, Alexander and Balog, Matej and Kumar, M. Pawan and Dupont, Emilien and Ruiz, Francisco J. R. and Ellenberg, Jordan S. and Wang, Pengming and Fawzi, Omar and Kohli, Pushmeet and Fawzi, Alhussein},
  year={2023},
  month=Dec,
  pages={468–475}
}

@inproceedings{lange2025shinkaevolve,
title={ShinkaEvolve: Towards Open-Ended and Sample-Efficient Program Evolution},
author={Robert Tjarko Lange and Yuki Imajuku and Edoardo Cetin},
booktitle={The Fourteenth International Conference on Learning Representations},
year={2026},
url={https://openreview.net/forum?id=lKEdGCoDNC}
}

@software{sharma2025openevolve,
  title = {OpenEvolve: an open-source evolutionary coding agent},
  author = {Asankhaya Sharma},
  year = {2025},
  publisher = {GitHub},
  url = {https://github.com/algorithmicsuperintelligence/openevolve}
}

@inproceedings{liu2026evox,
title={EvoX: Meta-Evolution for Automated Discovery},
author={Shu Liu and Shubham Agarwal and Monishwaran Maheswaran and Mert Cemri and Qiuyang Mang and Zhifei Li and Ashwin Naren and Ethan Boneh and Audrey Cheng and Alexander Du and Melissa Pan and Kurt Keutzer and Alvin Cheung and Koushik Sen and Alex Dimakis and Matei Zaharia and Ion Stoica},
booktitle={Third Conference on Language Modeling},
year={2026},
url={https://openreview.net/forum?id=puRQNPwKs1}
}

@inproceedings{ye2024reevo,
 author = {Ye, Haoran and Wang, Jiarui and Cao, Zhiguang and Berto, Federico and Hua, Chuanbo and Kim, Haeyeon and Park, Jinkyoo and Song, Guojie},
 booktitle = {Advances in Neural Information Processing Systems},
 doi = {10.52202/079017-1381},
 editor = {A. Globerson and L. Mackey and D. Belgrave and A. Fan and U. Paquet and J. Tomczak and C. Zhang},
 pages = {43571--43608},
 publisher = {Curran Associates, Inc.},
 title = {ReEvo: Large Language Models as Hyper-Heuristics with Reflective Evolution},
 url = {https://proceedings.neurips.cc/paper_files/paper/2024/file/4ced59d480e07d290b6f29fc8798f195-Paper-Conference.pdf},
 volume = {37},
 year = {2024}
}

@InProceedings{liu2024eoh,
  title = 	 {Evolution of Heuristics: Towards Efficient Automatic Algorithm Design Using Large Language Model},
  author =       {Liu, Fei and Xialiang, Tong and Yuan, Mingxuan and Lin, Xi and Luo, Fu and Wang, Zhenkun and Lu, Zhichao and Zhang, Qingfu},
  booktitle = 	 {Proceedings of the 41st International Conference on Machine Learning},
  pages = 	 {32201--32223},
  year = 	 {2024},
  editor = 	 {Salakhutdinov, Ruslan and Kolter, Zico and Heller, Katherine and Weller, Adrian and Oliver, Nuria and Scarlett, Jonathan and Berkenkamp, Felix},
  volume = 	 {235},
  series = 	 {Proceedings of Machine Learning Research},
  month = 	 {21--27 Jul},
  publisher =    {PMLR},
  url = 	 {https://proceedings.mlr.press/v235/liu24bs.html}
}

@misc{cemri2026adaevolve,
      title={AdaEvolve: Adaptive LLM Driven Zeroth-Order Optimization}, 
      author={Mert Cemri and Shubham Agrawal and Akshat Gupta and Shu Liu and Audrey Cheng and Qiuyang Mang and Ashwin Naren and Lutfi Eren Erdogan and Koushik Sen and Matei Zaharia and Alex Dimakis and Ion Stoica},
      year={2026},
      eprint={2602.20133},
      archivePrefix={arXiv},
      primaryClass={cs.NE},
      url={https://arxiv.org/abs/2602.20133}, 
}

@inproceedings{jiang2026deltaevolve,
title={DeltaEvolve: Accelerating Scientific Discovery through Momentum-Driven Evolution},
author={Jiachen Jiang and Tianyu Ding and Zhihui Zhu},
booktitle={Forty-third International Conference on Machine Learning},
year={2026},
url={https://openreview.net/forum?id=ownPRHrXMx}
}

@misc{yan2026pacevolve,
      title={PACEvolve: Enabling Progress-Aware Consistent Evolution}, 
      author={Minghao Yan and Bo Peng and Benjamin Coleman and Ziqi Chen and Zhouhang Xie and Shuo Chen and Zhankui He and Noveen Sachdeva and Isabella Ye and Weili Wang and Chi Wang and Ed H. Chi and Fernando Pereira and Wang-Cheng Kang and Derek Zhiyuan Cheng and Beidou Wang},
      year={2026},
      eprint={2601.10657},
      archivePrefix={arXiv},
      primaryClass={cs.NE},
      url={https://arxiv.org/abs/2601.10657}, 
}

@misc{assumpcao2025codeevolve,
      title={CodeEvolve: an open source evolutionary coding agent for algorithmic discovery and optimization}, 
      author={Henrique Assumpção and Diego Ferreira and Leandro Campos and Fabricio Murai},
      year={2026},
      eprint={2510.14150},
      archivePrefix={arXiv},
      primaryClass={cs.AI},
      url={https://arxiv.org/abs/2510.14150}, 
}

@misc{lehman2022elm,
      title={Evolution through Large Models}, 
      author={Joel Lehman and Jonathan Gordon and Shawn Jain and Kamal Ndousse and Cathy Yeh and Kenneth O. Stanley},
      year={2022},
      eprint={2206.08896},
      archivePrefix={arXiv},
      primaryClass={cs.NE},
      url={https://arxiv.org/abs/2206.08896}, 
}

@inproceedings{zhang2025dgm,
title={Darwin G"odel Machine: Open-Ended Evolution of Self-Improving Agents},
author={Jenny Zhang and Shengran Hu and Cong Lu and Robert Tjarko Lange and Jeff Clune},
booktitle={The Fourteenth International Conference on Learning Representations},
year={2026},
url={https://openreview.net/forum?id=pUpzQZTvGY}
}

@inproceedings{zelikman2023stop,
title={Self-Taught Optimizer ({STOP}): Recursively Self-Improving Code Generation},
author={Eric Zelikman and Eliana Lorch and Lester Mackey and Adam Tauman Kalai},
booktitle={First Conference on Language Modeling},
year={2024},
url={https://openreview.net/forum?id=46Zgqo4QIU}
}

@misc{zheng2026dreamrsi,
      title={Dream-RSI: Recursive Self-Improvement through Evolving Worlds}, 
      author={Tong Zheng and Xidong Wu and Zheng Zhang and Zhankui He and Chaoyi Zhang and Benjamin Coleman and Ruoqiao Wei and Di Bai and Haolin Liu and Rui Liu and Xue Wang and Yue Zhuan and Wang-Cheng Kang and Renkai Xiang and Heng Huang and Xinwu Cheng and Yunsong Guo},
      year={2026},
      eprint={2609.14858},
      archivePrefix={arXiv},
      primaryClass={cs.CL},
      url={https://arxiv.org/abs/2609.14858}, 
}

@misc{zhang2026hyperagents,
      title={Hyperagents}, 
      author={Jenny Zhang and Bingchen Zhao and Wannan Yang and Jakob Foerster and Jeff Clune and Minqi Jiang and Sam Devlin and Tatiana Shavrina},
      year={2026},
      eprint={2603.19461},
      archivePrefix={arXiv},
      primaryClass={cs.AI},
      url={https://arxiv.org/abs/2603.19461}, 
}

@misc{xia2026rrsi,
      title={RRSI: Regularized Recursive Self-Improvement of Agent Harnesses}, 
    author={Peng Xia and Rujun Han and Zifeng Wang and Yanfei Chen and Yufan Zhuang and Yoonho Lee and Chengsong Huang and Han Yu and Zhongying CuiZhu and Yifei Ming and Huaxiu Yao and Burak Gokturk and Tomas Pfister and Chen-Yu Lee},
      year={2026},
      eprint={2609.24972},
      archivePrefix={arXiv},
      primaryClass={cs.LG},
      url={https://arxiv.org/abs/2609.24972}, 
}

@misc{lee2026metaharness,
      title={Meta-Harness: End-to-End Optimization of Model Harnesses}, 
      author={Yoonho Lee and Roshen Nair and Qizheng Zhang and Kangwook Lee and Omar Khattab and Chelsea Finn},
      year={2026},
      eprint={2603.28052},
      archivePrefix={arXiv},
      primaryClass={cs.AI},
      url={https://arxiv.org/abs/2603.28052}, 
}

@misc{lee2026rhi,
      title={Recursive Harness Self-Improvement}, 
      author={Hyunin Lee and Jinglue Xu and Jeffrey Seely and Donghyun Lee and Matei Zaharia and Yujin Tang},
      year={2026},
      eprint={2607.15524},
      archivePrefix={arXiv},
      primaryClass={cs.LG},
      url={https://arxiv.org/abs/2607.15524}, 
}

@misc{lin2026ahe,
      title={Agentic Harness Engineering: Observability-Driven Automatic Evolution of Coding-Agent Harnesses}, 
      author={Jiahang Lin and Shichun Liu and Chengjun Pan and Lizhi Lin and Shihan Dou and Zhiheng Xi and Xuanjing Huang and Hang Yan and Zhenhua Han and Tao Gui and Yu-Gang Jiang},
      year={2026},
      eprint={2604.25850},
      archivePrefix={arXiv},
      primaryClass={cs.CL},
      url={https://arxiv.org/abs/2604.25850}, 
}

@misc{liu2026solpi,
      title={SoL-Pi: Recursively Scaling Auto-Research Loops for Efficient Agent Harness}, 
      author={Haozhe Liu and Tian Ye and Sensen Gao and Qihang Cao and Yitong Li and Mingchen Zhuge and Duomin Wang and Ruihua Zhang and Ping Luo and Jiawang Bian and Lei Zhu and Ligeng Zhu and Enze Xie and Song Han},
      year={2026},
      eprint={2609.20519},
      archivePrefix={arXiv},
      primaryClass={cs.AI},
      url={https://arxiv.org/abs/2609.20519}, 
}

@misc{srikanth2026aide2,
      title={Recursive Self-Improvement of AI Research Agents}, 
      author={Dhruv Srikanth and Bingchen Zhao and Dixing Xu and Yuxiang Wu and Zhengyao Jiang},
      year={2026},
      eprint={2609.26457},
      archivePrefix={arXiv},
      primaryClass={cs.AI},
      url={https://arxiv.org/abs/2609.26457}, 
}

@misc{fu2026sift,
      title={Self Improvement via Fast Tree-search}, 
      author={Xinghong Fu and Aravinth Kulanthaivelu and Yutaro Yamada},
      year={2026},
      eprint={2609.19526},
      archivePrefix={arXiv},
      primaryClass={cs.AI},
      url={https://arxiv.org/abs/2609.19526}, 
}

@inproceedings{liu2026skydiscover,
author = {Liu, Shu and Cemri, Mert and Agarwal, Shubham and Krentsel, Alexander and Naren, Ashwin and Mang, Qiuyang and Li, Zhifei and Gupta, Akshat and Maheswaran, Monishwaran and Cheng, Audrey and Pan, Melissa and Boneh, Ethan and Ramchandran, Kannan and Sen, Koushik and Zaharia, Matei and Dimakis, Alexandros G. and Stoica, Ion},
title = {SkyDiscover: A Flexible, Adaptive Framework for AI-Driven Scientific and Algorithmic Discovery},
year = {2026},
isbn = {9798400724152},
publisher = {Association for Computing Machinery},
address = {New York, NY, USA},
url = {https://doi.org/10.1145/3786335.3813221},
doi = {10.1145/3786335.3813221},
booktitle = {Proceedings of the ACM Conference on AI and Agentic Systems},
pages = {1223–1227},
numpages = {5},
location = {San Jose, CA, USA},
series = {CAIS '26}
}

@misc{imajuku2025alebench,
      title={ALE-Bench: A Benchmark for Long-Horizon Objective-Driven Algorithm Engineering}, 
      author={Yuki Imajuku and Kohki Horie and Yoichi Iwata and Kensho Aoki and Naohiro Takahashi and Takuya Akiba},
      year={2025},
      eprint={2506.09050},
      archivePrefix={arXiv},
      primaryClass={cs.AI},
      url={https://arxiv.org/abs/2506.09050}, 
}

@InProceedings{ouyang2025kernelbench,
  title = 	 {{K}ernel{B}ench: Can {LLM}s Write Efficient {GPU} Kernels?},
  author =       {Ouyang, Anne and Guo, Simon and Arora, Simran and Zhang, Alex L and Hu, William and Re, Christopher and Mirhoseini, Azalia},
  booktitle = 	 {Proceedings of the 42nd International Conference on Machine Learning},
  pages = 	 {47356--47415},
  year = 	 {2025},
  editor = 	 {Singh, Aarti and Fazel, Maryam and Hsu, Daniel and Lacoste-Julien, Simon and Berkenkamp, Felix and Maharaj, Tegan and Wagstaff, Kiri and Zhu, Jerry},
  volume = 	 {267},
  series = 	 {Proceedings of Machine Learning Research},
  month = 	 {13--19 Jul},
  publisher =    {PMLR},
  url = 	 {https://proceedings.mlr.press/v267/ouyang25a.html}
}

@misc{autokernel,
      title={AutoKernel: Autonomous GPU Kernel Optimization via Iterative Agent-Driven Search}, 
      author={Jaber Jaber and Osama Jaber},
      year={2026},
      eprint={2603.21331},
      archivePrefix={arXiv},
      primaryClass={cs.LG},
      url={https://arxiv.org/abs/2603.21331}, 
}

@misc{SwarmResearch,
      title={SwarmResearch: Orchestrating Coding Agents for Open-Ended Discovery}, 
      author={Yuvraj Virk and Zack Edds and Chunqiu Steven Xia and Lingming Zhang},
      year={2026},
      eprint={2607.02807},
      archivePrefix={arXiv},
      primaryClass={cs.AI},
      url={https://arxiv.org/abs/2607.02807}, 
}

@misc{aiscientist,
      title={The AI Scientist: Towards Fully Automated Open-Ended Scientific Discovery}, 
      author={Chris Lu and Cong Lu and Robert Tjarko Lange and Jakob Foerster and Jeff Clune and David Ha},
      year={2024},
      eprint={2408.06292},
      archivePrefix={arXiv},
      primaryClass={cs.AI},
      url={https://arxiv.org/abs/2408.06292}, 
}

@article{Gottweis_2026,
   title={Accelerating scientific discovery with Co-Scientist},
   volume={655},
   ISSN={1476-4687},
   url={http://dx.doi.org/10.1038/s41586-026-10644-y},
   DOI={10.1038/s41586-026-10644-y},
   number={8122},
   journal={Nature},
   publisher={Springer Science and Business Media LLC},
   author={Gottweis, Juraj and Weng, Wei-Hung and Daryin, Alexander and Tu, Tao and Sirkovic, Petar and Myaskovsky, Artiom and Glowaty, Grzegorz and Weissenberger, Felix and Orlandi, Alessio and Popovici, Dan and Palepu, Anil and Rong, Keran and Tanno, Ryutaro and Saab, Khaled and Zhang, Fan and Blum, Jacob and Carroll, Andrew and Kulkarni, Kavita and Tomašev, Nenad and Zverinski, Dina and Rendulic, Ivor and Vedadi, Elahe and Hasler, Florian and Rimanic, Luka and Boia, Marina and Budiselic, Ivan and Feinstein, Ben and Bellaiche, Mathias and Sheffer, Tom and Freyberg, Jan and Ratcliff, Jeremy and Bertolli, Ottavia and Chou, Katherine and Hassidim, Avinatan and Gokturk, Burak and Vahdat, Amin and Guan, Yuan and Dhillon, Vikram and Vaishnav, Eeshit Dhaval and Lee, Byron and Costa, Tiago R. D. and Penadés, José R. and Peltz, Gary and Matias, Yossi and Manyika, James and Hassabis, Demis and Xu, Yunhan and Kohli, Pushmeet and Pawlosky, Annalisa and Karthikesalingam, Alan and Natarajan, Vivek},
   year={2026},
   month=May,
   pages={487–496}
}

@inproceedings{coral,
title={{CORAL}: Towards Autonomous Multi-Agent Evolution for Open-Ended Discovery},
author={Ao Qu and Han Zheng and Zijian Zhou and Yihao Yan and Yihong Tang and Shao Yong Ong and Fenglu Hong and Kaichen Zhou and Chonghe Jiang and Minwei Kong and Jiacheng Zhu and Sirui Li and Cathy Wu and Bryan Kian Hsiang Low and Jinhua Zhao and Paul Pu Liang},
booktitle={Third Conference on Language Modeling},
year={2026},
url={https://openreview.net/forum?id=QcAiEdP4aF}
}

@misc{deepseek2026harness,
  author = {{DeepSeek-AI}},
  title = {{DeepSeek Harness}},
  year = {2026},
  howpublished = {\url{https://github.com/deepseek-ai/deepseek-harness}},
  url = {https://github.com/deepseek-ai/deepseek-harness}
}

@misc{cheng2025barbarians,
      title={Barbarians at the Gate: How AI is Upending Systems Research}, 
author={Audrey Cheng and Shu Liu and Melissa Pan and Zhifei Li and Bowen Wang and Alex Krentsel and Tian Xia and Mert Cemri and Jongseok Park and Shuo Yang and Jeff Chen and Lakshya Agrawal and Aditya Desai and Jiarong Xing and Koushik Sen and Matei Zaharia and Ion Stoica},
      year={2025},
      eprint={2510.06189},
      archivePrefix={arXiv},
      primaryClass={cs.AI},
      url={https://arxiv.org/abs/2510.06189}, 
}

@misc{shi2026composability,
      title={A Programming Paradigm for Spatiotemporal Composability}, 
      author={Yifan Shi and Wei Zhang and Tianyi Cui},
      year={2026},
      eprint={2608.25512},
      archivePrefix={arXiv},
      primaryClass={cs.PL},
      url={https://arxiv.org/abs/2608.25512}, 
}

@misc{googledeepmind2025alphaevolve_results,
  author = {{Google DeepMind}},
  title = {{Results of AlphaEvolve: Mathematical Results}},
  year = {2025},
  howpublished = {\url{https://github.com/google-deepmind/alphaevolve_results/blob/main/mathematical_results.ipynb}},
  url = {https://github.com/google-deepmind/alphaevolve_results/blob/main/mathematical_results.ipynb}
}

@misc{karpathy2026autoresearch,
  author       = {Andrej Karpathy},
  title        = {autoresearch: {AI} Agents Running Research on Single-GPU Nanochat Training Automatically},
  year         = {2026},
  howpublished = {\url{https://github.com/karpathy/autoresearch}},
  url          = {https://github.com/karpathy/autoresearch},
  note         = {GitHub repository}
}

@misc{lutke2026liquid,
  author       = {Tobias L{\"u}tke},
  title        = {Performance: 53\% Faster Parse+Render, 61\% Fewer Allocations},
  year         = {2026},
  month        = mar,
  howpublished = {GitHub Pull Request},
  note         = {Shopify Liquid, Pull Request \#2056},
  url          = {https://github.com/Shopify/liquid/pull/2056}
}

@misc{artificialanalysis2026deepseek,
  author = {{Artificial Analysis}},
  title = {{Model Comparison: DeepSeek V4 Pro and DeepSeek V4.1 Flash}},
  year = {2026},
  howpublished = {\url{https://artificialanalysis.ai/models/releases/comparisons?compare=deepseek-v4-pro,deepseek-v4-1-flash}},
  url = {https://artificialanalysis.ai/models/releases/comparisons?compare=deepseek-v4-pro,deepseek-v4-1-flash},
  note = {Accessed 2026-10-13}
}
